\documentclass[lettersize,journal]{IEEEtran}
\usepackage{amsmath,amsfonts}
\usepackage{algorithm}
\usepackage{array}
\usepackage[caption=false,font=normalsize,labelfont=sf,textfont=sf]{subfig}
\usepackage{textcomp}
\usepackage{stfloats}
\usepackage{url}
\usepackage{varwidth}
\usepackage{verbatim}
\usepackage{graphicx}
\usepackage{cite}
\usepackage[numbers,sort&compress,square]{natbib}
\usepackage{booktabs}       
\usepackage{amsfonts}       
\usepackage{nicefrac}       
\usepackage{microtype}      
\usepackage{xcolor}         
\usepackage{tablefootnote}
\usepackage{mathtools}
\usepackage{amsmath}
\usepackage{graphicx} 
\usepackage[flushleft]{threeparttable}
\usepackage{multirow}

\usepackage{amsmath}
\usepackage{amssymb}

\usepackage{algpseudocode}

\DeclarePairedDelimiterX{\infdivx}[2]{(}{)}{%
  #1\;\delimsize\|\;#2%
}
\newcommand{\infdiv}{KL\infdivx}

\usepackage[T1]{fontenc}

\begin{document}

\title{Automating the Learning of Inverse Kinematics for Robotic Arms with Redundant DoFs}

\author{Chi-Kai Ho, Chung-Ta King
\thanks{Chi-Kai Ho and Chung-Ta King are with the Department of Computer Science, National Tsing Hua University, Hsinchu, Taiwan. E-mail: s104062580@m104.nthu.edu.tw, king@cs.nthu.edu.tw}}

\markboth{Preprint,~Vol.~x, No.~x, August~2022}%
{Shell \MakeLowercase{\textit{et al.}}: A Sample Article Using IEEEtran.cls for IEEE Journals}


\maketitle

\begin{abstract}
Inverse Kinematics (IK) solves the problem of mapping from the Cartesian space to the joint configuration space of a robotic arm. It has a wide range of applications in areas such as computer graphics, protein structure prediction, and robotics. With the vast advances of artificial neural networks (NNs), many researchers recently turned to data-driven approaches to solving the IK problem. Unfortunately, NNs become inadequate for robotic arms with redundant Degrees-of-Freedom (DoFs). This is because such arms may have multiple angle solutions to reach the same desired pose, while typical NNs only implement one-to-one mapping functions, which associate just one consistent output for a given input. In order to train usable NNs to solve the IK problem, most existing works employ customized training datasets, in which every desired pose only has one angle solution. This inevitably limits the generalization and automation of the proposed approaches. This paper breaks through at two fronts: (1) a systematic and mechanical approach to training data collection that covers the entire working space of the robotic arm, and can be fully automated and done only once after the arm is developed; and (2) a novel NN-based framework that can leverage the redundant DoFs to produce multiple angle solutions to any given desired pose of the robotic arm. The latter is especially useful for robotic applications such as obstacle avoidance and posture imitation.
\end{abstract}

\maketitle

\section{Introduction}
\label{intro}
\IEEEPARstart{A} key computation in robot control is solving the \emph{inverse kinematics} (IK) problem. For a robotic arm, the problem is to find the \emph{angle solution}, i.e., the angle of each of the joint, that can move the end effector of the arm to a desired \emph{pose}, which consists of a target position and an optional orientation. The IK problem has been studied extensively since the 1980s~\citep{tolani2000real,neppalli2009closed,asfour2003human,zaplana2018novel,
ali2010closed,wampler1986manipulator,nakamura1986inverse,buss2005selectively}, and has found applications in such diverse areas as computer graphics, protein structure prediction, robotics, etc. The problem is commonly solved analytically based on the structure of the robotic arm~\citep{tolani2000real,neppalli2009closed,asfour2003human,zaplana2018novel,ali2010closed} or numerically by approximation through iterative calculations~\citep{wampler1986manipulator,nakamura1986inverse,buss2005selectively}, e.g., Jacobian-based methods.

With the vast advances of artificial neural networks, many researchers recently turned to data-driven approaches to solving the IK problem~\citep{duka2014neural,csiszar2017solving, almusawi2016new, ghasemi2019kinematic, srisuk2017inverse, khaleel2018inverse, toquica2021analytical, demby2019study, masuda2019common, polyzos2019solving}. Data-driven approaches can accelerate the deployment of robot applications~\citep{csiszar2017solving}, e.g., by simplifying calibration and not requiring deep expertises to apply, and can overcome problems such as singular configurations~\citep{almusawi2016new}. Most data-driven methods solve the IK problem using \emph{Deep Neural Networks} (DNNs)~\citep{duka2014neural,csiszar2017solving,almusawi2016new,ghasemi2019kinematic,srisuk2017inverse,khaleel2018inverse,toquica2021analytical,demby2019study}. DNN is a parameterized function, which can approximate a high-dimensional function for associating input with output by learning from data. To solve the IK problem, the neural network (NN) has to learn the mapping function from the desired pose to the angle solution.

The one-to-one mapping from the pose to the angle solution via the NN works fine for robotic arms without redundant \emph{Degree-of-Freedom} (DoF). However, it falls short for arms that have DoFs more than required, or \emph{redundant DoFs}. For example, if a 6-DoF robotic arm is only required to reach a target position without considering the orientation, then it has redundant DoFs to reach the target position with varying orientations. On the other hand, even if both position and orientation are given and required, a 7-DoF robotic arm still has an extra DoF to reach the given pose with different angle solutions, as shown in Fig. 1. For robotic arms with redundant DoFs, the one-to-one mapping function approximated by a NN cannot output multiple angle solutions given a desired pose as the input. If a given input has multiple targets to map, then those outputs are in competition for updating the weights of the NN while training, causing the training to fail. A common strategy to solving multiple outputs in DNNs is to concatenate all the outputs in a sequence and consider the sequence as a single output. However, such a strategy is not suitable here, because robotic arms with redundant DoFs can have an infinite number of angle solutions. It is not possible to concatenate them all together.

Most existing works on using NNs to solve IK work around the one-to-one mapping problem by considering specific applications with customized training datasets, in which every desired pose only has one angle solution. The problem is that the customized datasets normally do not cover the whole working space of the robotic arm. New data have to be collected and the NN model has to be retrained if the application is changed or the trajectories are altered. In addition, designing the customized training dataset for specific applications requires efforts and expertises, which cannot be automated easily. Collecting the training data is also a problem and often done manually, for example by manually pulling and pushing the robotic arm in the real world to the required poses and collecting the state of the actuators. Some people perform data collection in a virtual environment to speed up the process. The problem is that to move the robotic arm along certain designated trajectories may still involve traditional IK calculations and need expertises to a certain level. Finally, applications such as obstacle avoidance require multiple angle solutions to bypass the obstacles. Restricting the NN model to output only one angle solution is not desirable.

\begin{figure}
\centering
  \includegraphics[width=\linewidth]{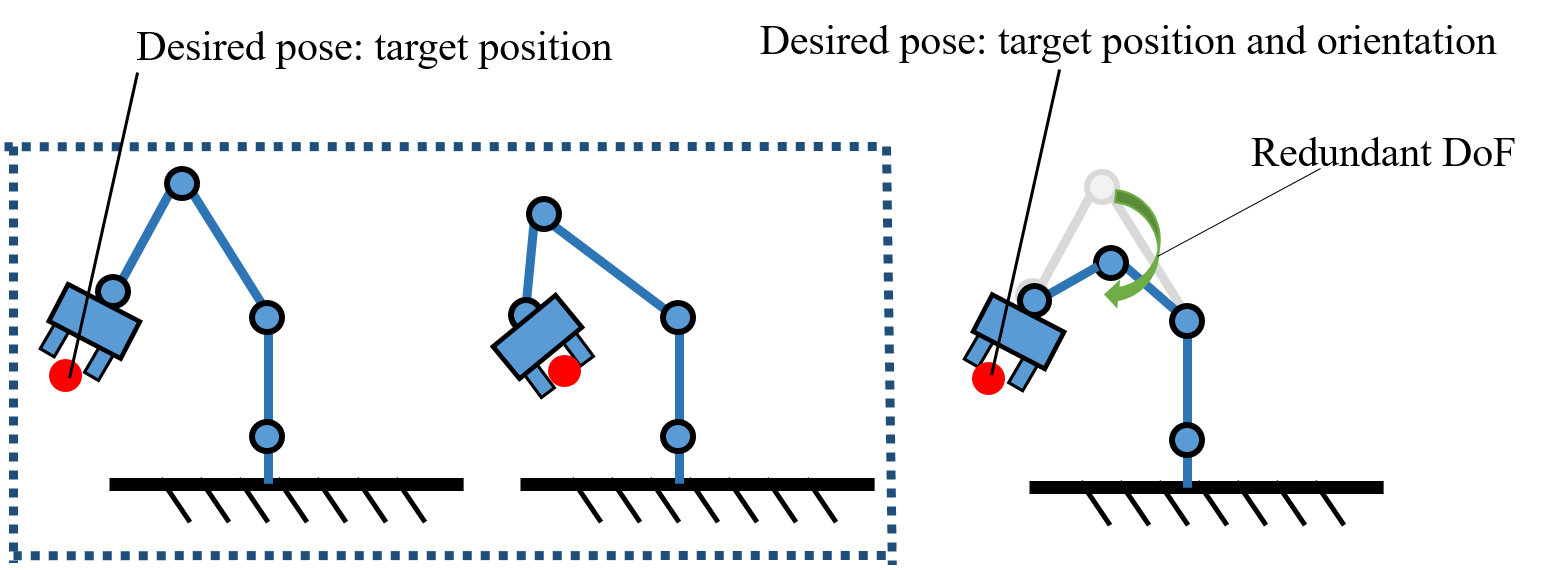}  
  \caption{An example of a robotic arm with multiple angle solutions when the desired pose consists of both position and orientation.}
  \label{overview}
\end{figure}

This paper presents the first comprehensive study on applying the data-driven approach to solving the IK problem for the entire working space of a robotic arm with redundant DoFs. We break through at two fronts: (1) a systematic approach to data collection and training for the entire working space that can be fully automated, generic for any application, and done only once after the robotic arm is designed; and (2) a novel NN model that allows multiple angle solutions to be generated for any given desired pose in the entire working space of the robotic arm. For the former, the basic idea is to systematically turn each of the joints of the robotic arm and record the position and orientation of the end effector. At the end, a dataset consisting of pairs of poses and angle solutions across the entire working space can be collected.  

\begin{figure}
\centering
  \includegraphics[width=\linewidth]{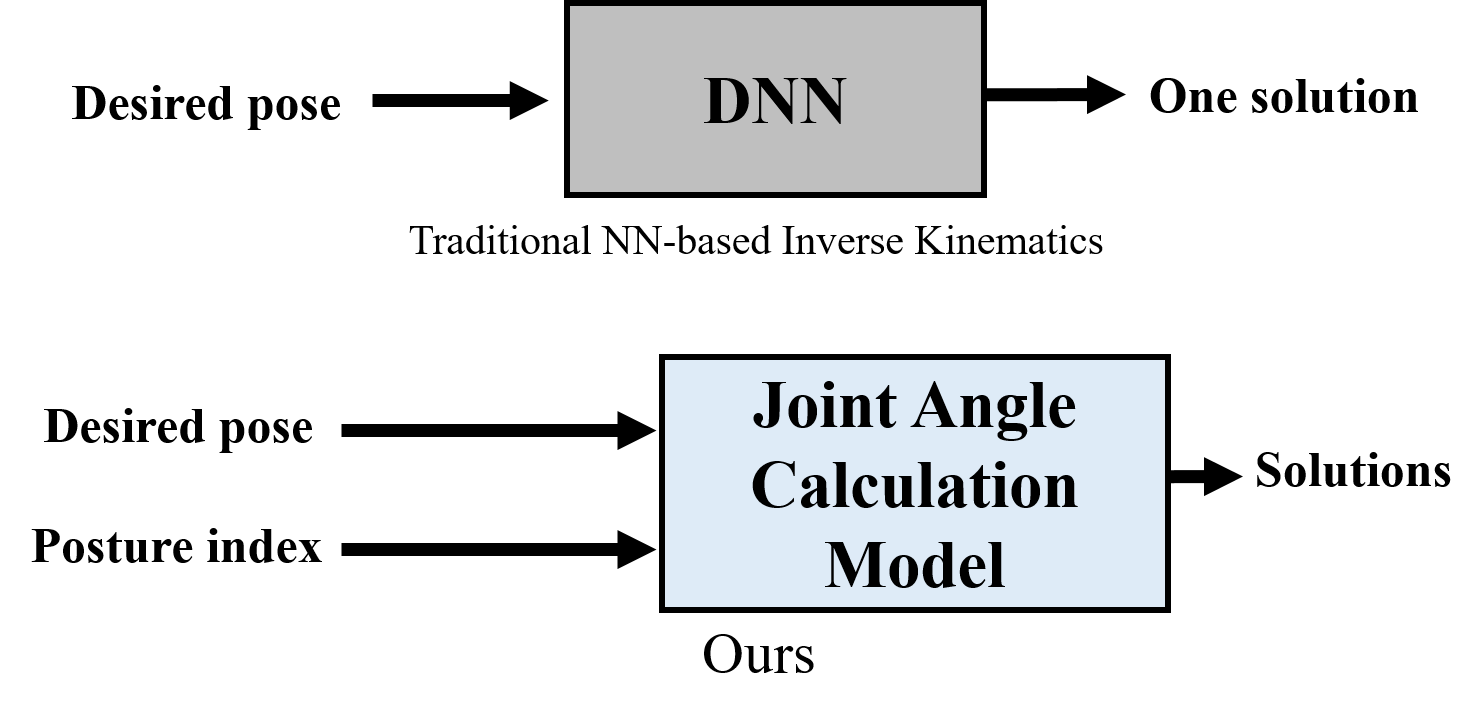}  
  \caption{Illustration of traditional DNN architectures (top) and our architecture (bottom).}
  \label{difference_tra_ours}
\end{figure}

The key insight to the latter is that an extra index, perhaps probabilistic, is needed to designate the varying angle solutions, as shown in Fig.~\ref{difference_tra_ours}. For reasons to be discussed later, we choose to learn a feature vector for representing the angle solutions for every possible pose of the robotic arm. The feature vector is referred to as the \emph{posture index}, which can be learned from the collected dataset in (1) through techniques such as \emph{Auto-Encoder} and \emph{Variational Auto-Encoder} (VAE) \citep{kingma2013auto,kingma2014semi,sohn2015learning}. As trained together with the posture indices, our NN model can easily associate the varying angle solutions in each pose with the rational feature vectors. What remained are tracking the posture indices for each possible pose for inference and representing the posture indices, e.g., discrete or probabilistic. We will discuss the different design options and present our proposed methods.

The main contributions of the paper are as follows: 
\begin{itemize}
  \item Given a required pose, the proposed method, called \emph{Selective Inverse Kinematics} (SIK), can find different angle solutions for a robotic arm with redundant DoFs by selecting one of the posture indices to generate the corresponding angle solution. To the best of our knowledge, this is the first work that solves the IK problem in the whole working space for robotic arms redundant DoFs.  
  \item The process to collect the training dataset can be fully automated, and the training of the NN model only needs to be done once after the arm is developed.
  \item The trained NN model is generic, covers the whole working space, and can be applied to any applications that require IK solutions.
  \item The NN model only requires very sparse training data to achieve a high accuracy in reaching the target position, to compare with the existing data-driven IK solutions. Experimental results show that the accuracy can be within ${\sim}$0.5 cm with training data collected by every $ 30^\circ $ per joint. 
  \item The proposed SIK has a fast computation time for real-time manipulations, and it can be applied to any robotic arm without hardware constraints.
  \item The paper provides a very comprehensive study of the proposed SIK, including properties of posture indices, effects of the sparsity of the training data, different training strategies, positions as well as orientations as the pose, etc.
\end{itemize}

The remainder of the paper is organized as follows. Section 2 introduces related works, and Section 3 describes the proposed SIK. Experiments and results are shown in Section 4. Conclusions are drawn in Section 5.

\section{Related Work}

Inverse kinematics (IK) calculates the angle displacement for each joint of a robotic arm such that the end-effector can reach the given target position with the given orientation in the working space. In other words, IK attempts to map from the robot working space (a Cartesian space) to its joint configuration space. Traditional solutions to the IK problem can be broadly categorized into two main classes: analytical (closed-form) methods \citep{tolani2000real,neppalli2009closed,asfour2003human,zaplana2018novel,ali2010closed} and numerical methods (Jacobian-based) \citep{wampler1986manipulator,nakamura1986inverse,buss2005selectively}.

Analytical methods use explicit mathematical formulations to solve the IK problem with closed-form expressions. They can precisely determine all possible IK solutions. The biggest shortcoming of analytical methods is the need to solve the algebraic formulas that are very complex and difficult. Another problem is that analytic IK solutions require full knowledge of the kinematic structure of the robotic arm, and so far the whole process is very difficult to automate. Numerical methods, on the other hand, approximate one IK solution through iterative approximation. They avoid the complex formulation process. However, they still require some basics of robotics, e.g., denavit-hartenberg parameters, forward kinematics, and pseudo-inverse Jacobian matrix.

In addition to the two traditional IK solutions, data-driven approaches~\citep{duka2014neural,csiszar2017solving, almusawi2016new, ghasemi2019kinematic, srisuk2017inverse, khaleel2018inverse, toquica2021analytical, demby2019study, masuda2019common, polyzos2019solving} have shown great potentials in solving the IK problem recently. Most of them applied a neural network to map the desired pose to an angle solution. They normally employed customized training datasets and could only infer desired poses in the designated area of the working space. Limited by the one-to-one mapping of the neural networks, existing data-driven methods can only find one angle solution for a given desired pose, which greatly restricts the applications of robotic arms with redundant DoFs.

In \citep{almusawi2016new}, an accurate solution for the IK problem by using a neural network is introduced. The main idea is that the neural network for solving the IK problem can be improved if the current state of the robotic arm is added to the input. Their training dataset can only have a unique joint configuration in both input and output sets because their model can only find one angle solution. The experiments showed high accuracy in robot motion control by following specific paths, but it is unclear whether the proposed method can infer in the entire working space. Compared with this work, we do not need a specialized training dataset. Our model can find the angle solutions for any desired pose in the entire working space. 

In \citep{csiszar2017solving}, a machine learning approach to solving the IK problem was proposed that can eliminate the need for developing the equations by hand. In \citep{srisuk2017inverse}, a work that can solve the IK of a 3-DoF robotic arm in 3-dimensional space was introduced, which does not need a large dataset for training and can reduce the neural network structure. In \citep{khaleel2018inverse}, an IK solution by combining Genetic Algorithm and Neural Network was proposed. The above three studies again solved the IK problem by providing only a single angle solution, and they need to put extra efforts into designing the customized training datasets, e.g., space-filling curves. Furthermore, these works only considered low-DoF robotic arms (2-DoF and 3-DoF), and it is unclear how they may be extended to redundant robotic arms. 

The work in \citep{ghasemi2019kinematic} introduced two approaches to the IK problem for the Tricept robot: one is based on the MLP neural network, and the other is based on the RBF neural network. The proposed approaches can only find one angle solution for a given pose and consider only low-DoF robots. The training data shown in this work is a limited convex set. The study in \citep{demby2019study} discussed solving the IK problem of multiple robotic arms using Artificial Neural Networks and Adaptive Neuro-Fuzzy Inference Systems. Similarly, this work adopted customized structured workspaces to sample thousand data points, so the available testing range is bounded and can only obtain one angle solution for a target position. Although it experimented with 4, 5, 6, and 7-DoF robotics, the accuracy is poor, over a few centimeters. By contrast, our approach has a low distance error (smaller than 0.5 cm) and can infer poses in the whole working space. 

Two solutions were proposed in \citep{toquica2021analytical} for the IK problem of an industrial parallel robot: a closed analytical form and a deep-learning approximation model based on three different networks (MLP, LSTM, GRU). The training data was collected through a user-defined point cloud and a 3-DoF parallel robot, the IRB360, is used for experiments. Again, the work ignored the flexibility of redundant robotic arms by considering only the unique solutions confined by the point cloud. It is unclear how the proposed work performs for redundant robotic arms. 

In \citep{masuda2019common}, a new method for learning a mapping between redundant states and low-dimensional postures was proposed. The work considered a high-DoF, complex musculoskeletal robot and attempted to find sets of internal pressures of pneumatic artificial muscles to meet a target position. This work attempted to address the multiple-solution issue in robotic arms as well. They also exploited an auto-encoder to handle the training data and supervised learning to learn known low-dimensional corresponding vectors. To collect the training data, they move the arm first to the designated positions and randomly give different pressures to obtain multiple solutions. In contrast to their work, we focus more on the poses in the entire working space. Although their method can generate different solutions to the positions, the positions still need to be on the designated trajectories. The work lacked a comprehensive study of the learned vectors, and hence the impacts of the feature vectors were not evaluated. In this work, we show a comprehensive analysis of the posture indices, and our model can achieve very good performances in both speed and accuracy. The experimental results show that our model can achieve a position accuracy within 0.5 cm, which is much better than the method proposed in \citep{masuda2019common} (${\sim}$10 cm).

\section{Approach}
\label{headings}

\begin{figure}
\centering
  \includegraphics[width=\linewidth]{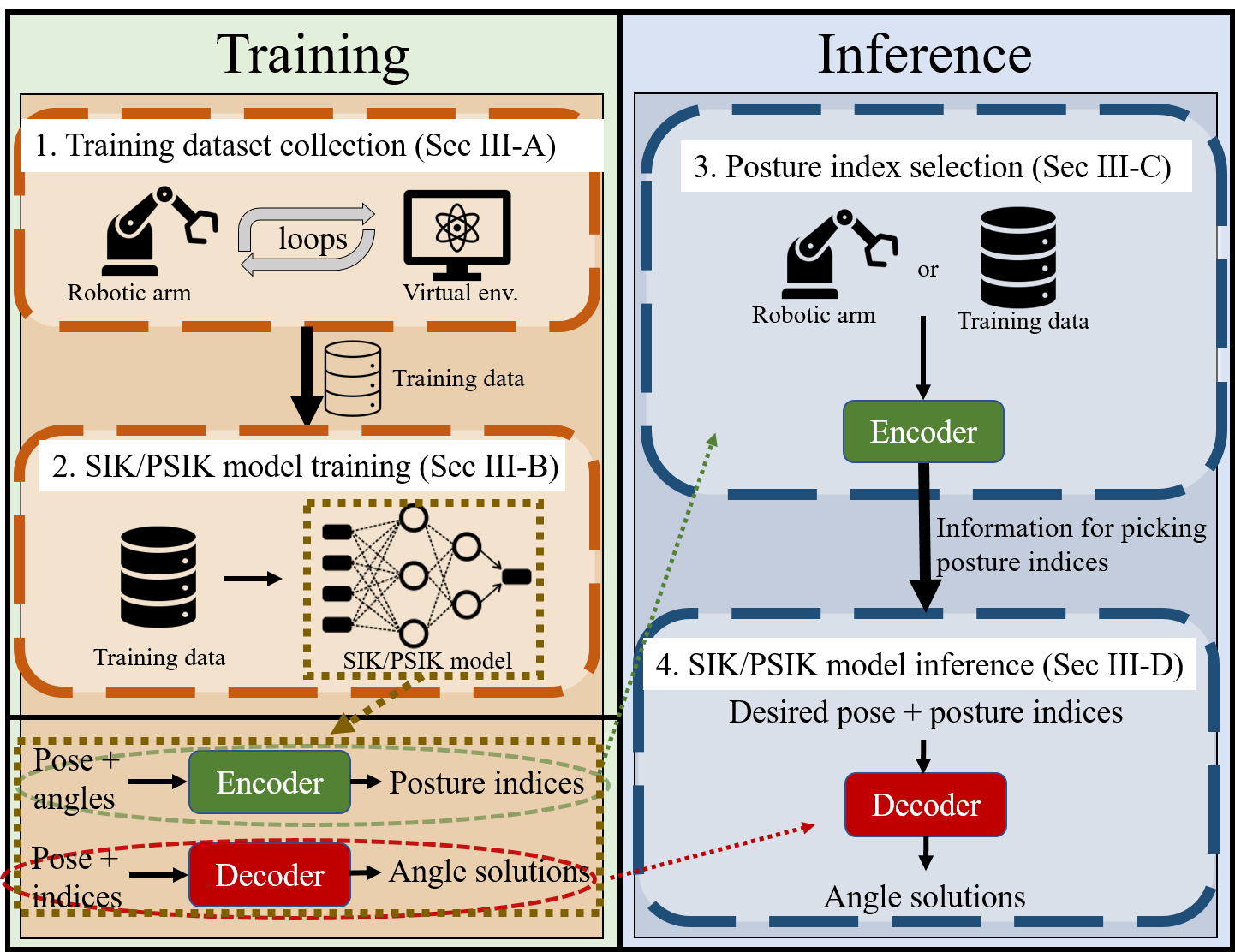}  
  \caption{An overview of the proposed SIK method.}
  \label{overview}
\end{figure}

Our overall goal is to simplify the process of data collection and provide a novel architecture that can supply multiple angle solutions at any pose in the whole working space for those arms with redundant DoFs. Fig.~\ref{overview} gives an overview of the proposed SIK method. In the training stage, we first collect an unbiased training dataset across the entire working space of the robotic arm. This step is easy and straightforward, which can be done with a few lines of code as described in Sec~\ref{method_trainingdata}. The collected dataset is a multi-modal dataset, in which a desire pose may have multiple angle solutions. Such a dataset cannot be used to train typical NN networks, which in essence are one-to-one mapping functions. To address this issue, we propose two novel NN architectures to associate a pose to its corresponding multiple angle solutions, which are presented in Sec~\ref{method_trainingSIK}. After the NN models are trained, we can apply the models during inference. Given a desired pose, the first step is to select the posture indices (see Sec~\ref{method_indexselection}). Next, the desired pose and the selected posture indices are fed into the trained model to produce the angle solution, as introduced in Sec~\ref{method_inference}.

\subsection{Unbiased training dataset collection} \label{method_trainingdata}

The proposed SIK method adopts a data-driven strategy to solving the IK problem. Therefore, the first step is to collect a training dataset for offline SIK neural network training. As mentioned earlier, most existing NN approaches to solving IK collect a biased training dataset by defining, for example, a point cloud or space-filling curves to construct a bounded region, in which there is only one angle solution for each desired pose. In this way, the NN models can be trained successively as one-to-one mapping functions. The problem is that the training dataset has to be specially and manually designed, which hinders the automation of the data collection process. Furthermore, if the applications or problem requirements are changed, new dataset has to be collected and the process has to be repeated. 

In this paper, since we aim to solve the one-to-many mapping from a desired pose to the multiple angle solutions for robotic arms with redundant DoFs, we can simply collect an unbiased multi-model dataset across the entire working space of the robotic arm. The idea is straightforward. Given a robotic arm, we turn each joint of the arm in turn by $x$ degrees across the working range of that joint. For each turn of a joint, we record the angles of all the joints and use a simulator to obtain the coordinates and orientations of the end-effector. This step can be accelerated if \emph{forward kinematics} (FK) is available to use. Algorithm~\ref{collect_data_algo} shows the collection flow in pseudocode. After all joints are turned across their working range in sequence, we should have collected an unbiased dataset that covers the entire working space of the robotic arm for all possible poses that the arm can perform, in which a pose may find multiple sets of angle configurations to reach it. This process only needs to be done once after the arm is designed. 

\begin{algorithm}
\small
\caption{Data collection (recursive version)} \label{collect_data_algo}
\begin{algorithmic}
\Require Joint upper limits: $upper\_limits$; Joint lower limits: $lower\_limits$; Joint interval for data collection: $interval\_x$; The degrees of freedom of the arm: $\#joint$;

\State
\Procedure{$data\_collection\_function$}{$\#joint$}
  \State
  \For{\texttt{$k \gets 0$ to $\#joint$}} \Comment{Initialize motors}
        \State $joint\_parameters[k] \gets lower\_limits[k]$
  \EndFor
   
  \State
  \If {$\#joint$ is 0} \Comment{Termination condition}
  \While{$joint\_parameters[0]  \leq upper\_limits[0]$}
  \State{Set the joints with $joint\_parameters$}
  \State{Record the pose of the end-effector}
  \State \begin{varwidth}[t]{0.6\linewidth} $joint\_parameters[0] \gets$\end{varwidth}
  \State \begin{varwidth}[t]{0.6\linewidth} $joint\_parameters[0] + interval\_x$\end{varwidth}
  \EndWhile
  \State{\textbf{Return} 0}
  \EndIf

  \State  
  \raggedright \While{$joint\_parameters[\#joint] <  upper\_limits[\#joint]$}
  \State{\textbf{Call} $data\_collection\_function$($\#joint$ - 1) }
  \State \begin{varwidth}[t]{0.8\linewidth} {$joint\_parameters[\#joint] \gets$} \end{varwidth}
  \State \begin{varwidth}[t]{0.8\linewidth} {$joint\_parameters[\#joint] + interval\_x$} \end{varwidth}
  \EndWhile
  
\EndProcedure

\end{algorithmic}
\end{algorithm}

\subsection{SIK/PSIK model training} \label{method_trainingSIK}

A typical NN model is a parameterized one-to-one mapping function, which calculates a corresponding and consistent output for a given input. If we use a multi-modal dataset, in which a pose has various solutions to map to, then it is not possible to find a set of parameters for the NN model to output all solutions at once, causing the training to fail.

One strategy to get multiple angle solutions given a desired pose is to extend the output layer of the NN by concatenating all the angle solutions of a pose for training and factorizing the long output during inference. The problem is that the angle solutions to a pose are countless, and thus it is impossible to define a bounded output size for the NN model. Another problem is that different poses can have quite different number of angle solutions. Usually, robotic arms have more angle solutions to reach the poses around the center of their working space than those near the boundary. It is thus necessary to solve the alignment problem of the angle solutions.

In this paper, we propose to modify the input layer of the NN model to accommodate the unbiased training dataset collected as discussed in Sec~\ref{method_trainingdata}. The idea is to add an extra index to the input layer, so the NN function maps from a given desired pose and an index to an angle solution. This solves the alignment problem, because the NN model only outputs a single angle solution at a time. Intuitively, we can simply use a sequence of integers as the index to associate different angle solutions. Unfortunately, this cannot be achieved in practice. The unbiased training data collected with the proposed method does not guarantee that the solutions are arranged well and evenly distributed, so these integer indices for adjacent poses sometimes are mapped to similar angle solutions, and other times they are not. Such stochastic labeling causes the network to fail in training. Hence, we need a better index that can have a consistent relationship to associate  angle solutions with different poses, which best match the characteristics of the solutions. 

Our idea in SIK is to learn an index, called \emph{posture index}, by unsupervised learning using the unbiased dataset collected in the previous section. The idea is inspired by \emph{Auto-encoder} \citep{bengio2009learning}, which is an unsupervised feature-learning scheme. In an auto-encoder, we use an encoder to compress desired poses and their angle solutions and a decoder to reconstruct the compressed data. By calculating the reconstruction loss, the posture indices are related to the angle solutions, and they have a strong and consistent meaning to the postures of the robotic arm. With these properties, SIK has a uniform form of labeling to associate different angle solutions for different poses.

We denote the \emph{desired goal}, which is a target position plus perhaps an orientation, of the robotic arm by \(\it{g}\). The set of angle solutions with respect to desired goal \(\it{g}\) is denoted by  \( \mathbb{A}_g = \{a_0^{g}, a_1^{g},\dots,a_n^{g}\} \), where \(a_i^{g}\) is a vector representing an angle solution for the given goal. Given the desired goal \(\it{g}\), ideally we want to find a function \( P(\mathbb{A}_g |g) \) to generate all angle solutions. However, obtaining the joint probability of \( \{a_0^{g}, a_1^{g},\dots,a_n^{g}\} \) is not realistic due to the infinite number of angle solutions of high-DoF robotic arms. Therefore, we simplify this problem by adopting an alternative function \(P(a_i^{g} |g,i_i^{g})\), where \(i_i^{g}\) is an index referring to one solution in the solution set \(\mathbb{A}_g\). With this function, it is possible to obtain different solutions in the set \(\mathbb{A}_g\) by providing different indices. Now, there is no need to calculate the solutions all at once, because the index can be used to generate individual solutions.

\begin{figure}
\centering
  \includegraphics[width=\linewidth]{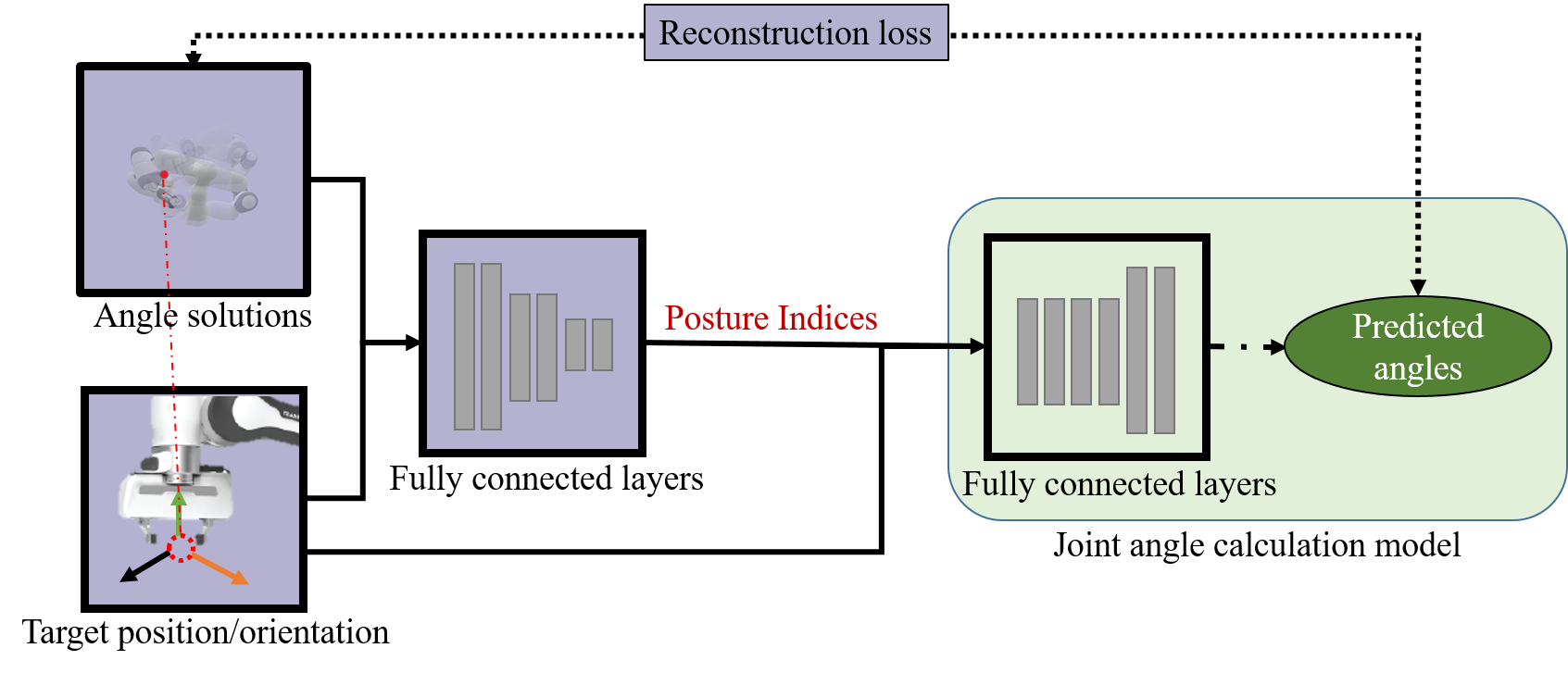}  
  \caption{The implementation of the SIK. }
  \label{implementation_autoencoder}
\end{figure}

Fig.~\ref{implementation_autoencoder} shows the implementation of our approach. The green part is our main angle calculation module for inference, and the rest of the architecture aims to obtain the posture indices. In the training phase, the encoder part first takes the poses and their corresponding angle solutions from the training dataset collected in Sec~\ref{method_trainingdata} and extracts feature vectors as the posture indices. The decoder part then receives a pose and its posture indices, deconverting this information back to the angle solutions by minimizing the reconstruction loss. After training, the decoder part in the figure can be used to calculate the angle solutions during inference.

Adopting Vanilla auto-encoder is sufficient for retrieving the characteristics of the postures, producing uniform labels. However, such a method does not consider the indices as continuous and may cause poor interpolations of the indices. Theoretically, an interpolated angle solution between two adjacent angle solutions should have a corresponding index between the indices of the two solutions. However, for each training iteration, Vanilla auto-encoder compresses an angle solution into some numbers without taking other solutions into account. Hence, it does not guarantee the posture indices have such property. To have more rational indices, we propose a variant of SIK, called \emph{PSIK}, which encodes the posture indices more rationally by mapping them to probability distributions. By introducing probabilistic distributions, the posture index becomes a small range instead of a point. In other words, now the posture indices have to not only make sure the correctness of the decoding but also concern the adjacent indices. 

\begin{figure}
\centering
  \includegraphics[width=\linewidth]{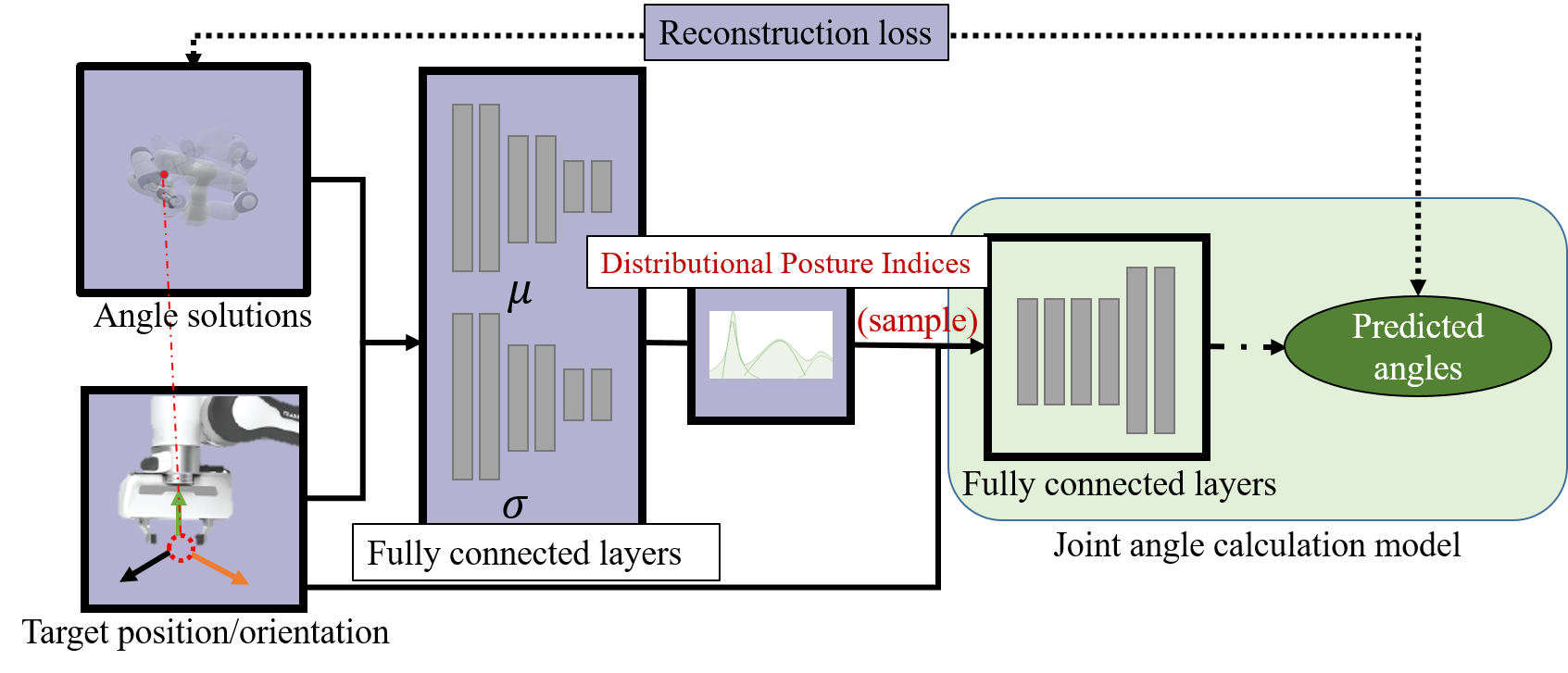}  
  \caption{The implementation of the PSIK. }
  \label{implementation_vae}
\end{figure}

To implement PSIK, we adopt \emph{Variational Auto-encoder} (VAE) \citep{kingma2013auto,kingma2014semi,sohn2015learning}. The VAE has the important advantage of approximating posterior with continuous latent variables. Fig.~\ref{implementation_vae} shows the architecture of PSIK. We first convert the target position and the angle solutions to a simple probability distribution, e.g., normal distribution, through an encoder. Then we sample latent variables, i.e., posture indices, from the distribution and convert the latent variables with the target position back to the angle solutions through a decoder. It should be noted that we do not really sample latent variables, because backpropagation cannot handle sampling. Therefore, the reparameterization technique is used to implement the idea~\citep{kingma2013auto}. The original equation in VAE~\citep{kingma2013auto} is modified slightly by adding extra conditions. The loss function is as follows:

\begin{equation*}
\begin{aligned}
 Loss = -\int_{z}^{} q(z|g, x)\cdot \log p(x|g, z) \,dz \\+ \infdiv{q(z|g, x)}{Nor(0, 1)},
\end{aligned}
\end{equation*}
 
\noindent where \(\it{g}\) is the target position, \(\it{x}\) is the corresponding angle solutions with respect to \(\it{g}\), and \(\it{z}\) is the posture index. The objective is to minimize the equation above.

\subsection{Posture index selection} \label{method_indexselection}

In Sec~\ref{method_trainingSIK}, we introduce the proposed architecture and how it associates a desired pose with multiple angle solutions. The architecture has an encoder for generating rational posture indices and a decoder for obtaining different angle solutions. During inference, the decoder receives the desired pose and the indices to generate the angle solutions. The problem is that, even though the posture indices indicate the angle solutions, every desired pose has its own set of angle solutions and thus a specific set of posture indices. In fact, the posture indices are more like indices to separated groups of similar postures of the robotic arm, e.g., postures that the elbow joint bends backward. Hence, we cannot casually apply some random number as the posture index of a desired pose and hope to obtain a usable angle solution. Different desired poses will have different groups to match and different sets of posture indices. 

One strategy to tracking the posture indices for different poses is to maintain a table to store poses and their corresponding posture indices. Although poses and posture indices are continuous across the working space of the robotic arm, the continuity in our NN models ensure that adjacent poses have similar angle solutions and posture indices. Therefore, a sparse and discrete table can serve the purpose. The question is which poses are to be stored in the table. There are various ways to do. In our implementation, we simply use the poses that are collected in the training dataset. The idea is that, after the encoder is trained, we feed all the poses collected in the training dataset to the encoder once more to obtain the corresponding posture indices. To reduce the amount of data stored in the table, we only store the indices with large differences and drop those that are too similar. We then use a  KD-tree and a dictionary data structure to store the posture indices of each pose that appears in the training dataset, for indicating the available ranges of the posture index for a given pose. During inference, given a desired pose, we search for the closest pose in the table and take its posture indices as the indices for the desired pose. In our experiments, we adopt this strategy, because it can easily obtain different solutions for the desired poses. We can then validate the performance of the models for various angle solutions.

Another strategy to obtaining usable posture indices for a desired pose is to leverage the current state of the arm, i.e., the joint configuration and the pose of the end-effector. Note that the current state signifies the initial position of the arm and the desired pose gives the target for the arm to move to. We can feed the current joint configuration and the desired pose into the encoder and obtain a posture index as the output. The implication is that the arm may move to the desired pose with a posture similar to that of the initial state of the arm. It is also possible to alter the posture index somewhat to obtain other angle solutions in the same posture group. Although this strategy cannot obtain all the angle solutions to the desired pose, it is sufficient to exploit alternative angle solutions to satisfy requirements such as obstacle avoidance. More concrete illustrations are shown in Tables~\ref{second_posture_index_table} and~\ref{vision-ori-table}. Furthermore, if the desired pose only requires a target position, we can transfer the current posture to another posture with the opposite orientation. We can also change the slope of the arm with redundant DoF without affecting the target position and the orientation. More discussions will be given in the next section.

\subsection{SIK/PSIK model inference} \label{method_inference}

Given a desired pose, we can apply the methods introduced in the previous subsection to obtain multiple posture indices for that pose. The posture indices can be fed together with the desired pose into the decoder, as shown in Fig.~\ref{overview}, to obtain the corresponding angle solutions to move the robotic arm to the desired pose. What remains is to determine which posture index is the best to use. Note that this may involve adjusting the posture indices produced directly from the methods in Sec~\ref{method_indexselection}.

The goodness of the posture indices really depends on the application requirements, which are often expressed as an evaluation or utility function. For example, if the application is for obstacle avoidance, then the evaluation function can use the arm's forward kinematics and distance algorithms to calculate the shortest distance of the arm to the obstacles and the minimum distance must be greater than a given threshold to keep the arm away from the obstacles. Algorithm~\ref{algo} shows a simple framework for evaluating and adjusting the posture indices. In the framework, we can define different evaluation functions for the posture indices according to the application requirements. The evaluation functions may include extra information, e.g., positions of the obstacles, and can determine how to adjust the posture index. In Algorithm~\ref{algo}, we show an example evaluation function, $\emph{F}(\theta, \emph{I})=\rho$, and demonstrate how each element of the posture index is adjusted in turn to find a posture index, which results in a pose that better meets the application requirements. 

Note that the proposed SIK method does not involve environmental information, nor does it establish any mathematical relationship between the posture index and the robot posture. However, as shown in Table~\ref{second_posture_index_table}, we can see that posture indices are continuous and can affect the posture of the arm gradually. Thus, we can add different values on the index to manipulate the robotic arm. The lack of environmental information in our model training is not a problem either. SIK focuses on controlling the robotic arm to move to the desired pose without involving any environmental information. Just like image-based robot control, in which imaging is processed before controls and separated from the control system, in SIK, environmental information can be handled by other systems, while SIK is responsible for obtaining the usable angle solutions according to the application requirements. In summary, this paper proposes a novel automation process to obtain multiple angle solutions to a desired pose for robotic arms with redundant DoFs. This opens up a new way to control robotic arms. The applications of robotic arms do not become limited when we change the way of operating them.
\begin{algorithm}
\caption{A unified framework for extended applications with the proposed method} \label{algo}
\begin{algorithmic}
\Require Joint angle calculate model: $\emph{J}(\{i_{1},,...i_{k}\}, pose)=\{\theta_{1}, \theta_{2},..., \theta_{n}\}$ where $i$ is the posture index and $\theta_{n}$ is the value of the $n$-th joint; An evaluate function: $\emph{F}(\theta, \emph{I})=\rho$, where $\rho$ indicates how compatible to the application the angle solution is and $\emph{I}$ is the extra required information; A required threshold of the compatibility: $\epsilon$;
\Ensure An angle solution that satisfies the requirements

\State
\State $\{i_{1},...i_{k}\}, \emph{I}, pose \gets Initialization$
\State $\{\theta_{1}, \theta_{2},..., \theta_{n}\} \gets \emph{J}(\{i_{1},...i_{k}\}, pose)$
\State $\rho \gets \emph{F}(\{\theta_{1}, \theta_{2},..., \theta_{n}\}, \emph{I})$
\State $\{x_{1}, x_{2},..., x_{k}\} \gets $find a vector that can increase the compatibility with evaluation function $\emph{F}$ when it is added into the posture index (can be done with brute-force algorithms)
\While{$\rho < \epsilon$}
	\State $\{i_{1},...i_{k}\} \gets \{i_{1}+x_{i},...i_{k}+x_{k}\}$
	\State $\{\theta_{1}, \theta_{2},..., \theta_{n}\} \gets \emph{J}(\{i_{1},...i_{k}\}, pose)$
	\State $\rho \gets \emph{F}(\{\theta_{1}, \theta_{2},..., \theta_{n}\}, \emph{I})$
\EndWhile
\end{algorithmic}
\end{algorithm}

\section{Experiments}

This section is organized as follows. In Section~\ref{ExpSetup}, details of the experimental setup are given. In Section~\ref{accuracy_performance}, we estimate the overall position accuracy of the proposed approach, and in Section~\ref{time_performance} the computation time of our approach is evaluated. In Section~\ref{various_solutions}, the diversity of the multiple angle solutions to reach a given target position is examined. In Section~\ref{section_posture_indices}, properties of the posture indices are studied. Section~\ref{orientation_performance} shows the results using the target position plus the orientation as the given pose. Finally, Section~\ref{extended_applications} demonstrates some applications of using the proposed method.

\subsection{Experimental Setup}\label{ExpSetup}

\begin{figure}
  \center
  \includegraphics[width=5cm]{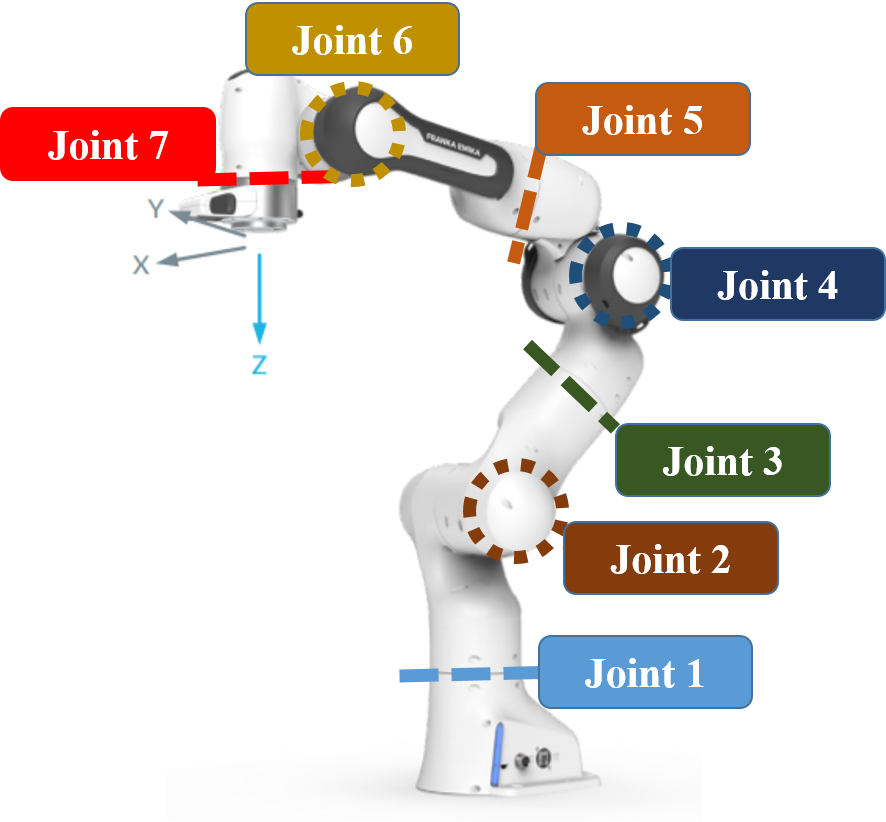}  
  \caption{The Franka Emika Panda robot is a 7-DoF robotic arm. The robot has a maximum stretch of 855 mm. Joint position limits (degrees) are as follows: A1, A3, A5, A7: -166/166, A2: -101/101, A4: -176/-4, A6: -1/215.}
  \label{arm_structure}
\end{figure}

We evaluate the proposed methods using a 7-DoF robotic arm, Franka Emika Panda Arm\footnote{https://www.franka.de/}. The robotic arm's structure is shown in Fig.~\ref{arm_structure}. The structure details are described below: Joint 1, which is the axis nearest the base, controls the orientation of the whole arm. Joint 4 is an elbow joint, which gives more versatility to the end-effector. The first three axes (Joints 1, 2, and 3) determine the position of the elbow joint, and the remaining axes determine the pose of the end-effector. The rotation limits of the joints are as follows: {Min/Max (degree)} = {A1: -166/166, A2: -101/101, A3: -166/166, A4: -176/-4, A5: -166/166, A6: -1/215, A7: -166/166}. The experiments were conducted entirely in a virtual environment, the PyBullet \citep{coumans2016pybullet} physics simulator, for avoiding unexpected collisions and damage. In the virtual environment, the physical parameters of the robot arm are the same as those of the real robot, except the self-collision mechanism was turned off. 

The proposed approach is compared against two other learning-based methods: an adaptive neuro-fuzzy inference system (ANFIS), which has been discussed in~\citep{demby2019study}, and a deep fully-connected neural network, as proposed in~\citep{demby2019study, toquica2021analytical}. Since we do not have any prior information and recommended values to set up the proposed network, and we do not have adequate hardware to finetune the hyperparameters, we applied a huge size of the neural network to avoid the problem of the lack of neurons. But it does not mean that the proposed approach must be trained with tremendous variables; in appendix B, we also show that the proposed method can be accomplished with a small number of neurons.

To implement SIK, the encoder and decoder each had five fully connected layers. The encoder had  2048, 2048, 1024, 512, and 4 neutrons in the five layers respectively, whereas the decoder had 512, 1024, 1024, 512, and 7 neutrons. Each layer was followed by an ELU activation function, except for the output layer. The posture index produced by the encoder consists of four floating-point numbers. We have tried different parameters to train the neural networks of SIK. It is found that the learning rate should be less than or equal to 0.0005, decreased after every 1000 training epochs, and the batch size should be 65536. Stable and satisfactory results can be obtained for the entire working space after 3300 epochs. In PSIK, the layers are almost the same as in SIK, but the output layer of the encoder generates two outputs: one stands for the mean, and the other stands for the variance.

The parameters of the ANFIS were as below: the number of premise functions of each feature was 2, the range of allowed values of the exponent in the premise functions was from 0.2 to 0.5, the range of allowed values of the exponent in the premise functions was from 1 to 3, and the range of allowed values of the coefficients in the consequent functions was from -10 to 10. The solver of the ANFIS was a particle swarm optimizer (PSO), in which the number of populations was 100 and the number of iterations was 200. Because of the limited memory space, ten premise functions each feature is the largest configuration we can have. Regarding the baseline, the deep fully-connected neural network had 3, 512, 1024, 1024, 512, and 7 neutrons in the five layers, which are identical to the decoder of our approach except for the input layer.

The training dataset consists of the motor angles of the arm, coordinates (x, y, z) of the end-effector, and the orientation of the end-effector. The dataset was collected every 30 degrees of each motor. The dataset contains 6,967,296 data points with a total size of 589.5 MB. From the collected data, we identified 580,608 different positions, resulting in about 12 data points per position on average. Note that it does not mean there are only 12 postures per position. Since the dataset is sparse, it is not possible to obtain all the angle solutions to the same target position. To cope with the sparsity and to generate a sufficient number of angle solutions to a target position for selection, we actually include all the data points that can reach within 1 cm from the given target position as the angle solutions to that position. Hence, in Section~\ref{various_solutions}, we also considered the posture indices from the neighboring nodes that are close to the target position for selection. In addition to this dataset, we also drop the data points that have the same target position of the end-effector to obtain the second dataset called the uni-position dataset. The second dataset was prepared for especially training the other two comparing methods. As stated previously, the data points with the same target positions (input) are considered noise while training a classic fully-connected network because the wights are in competition. Hence, we attempted to alleviate the impact of the "duplicated" data by removing those data with the same positions. The uni-position dataset contains 523,267 data points with a total size of 61.2 MB.

In the experimental section, we evaluate the proposed method from two different aspects: in the former part (from Section~\ref{accuracy_performance} to Section~\ref{time_performance}), we compare the proposed approach with other two methods to see the advantages of our method. The proposed methods are compared based on the accuracy of the end-effector in reaching the target position and the computation time. The accuracy is measured in terms of \emph{distance error}, which is defined as the distance from the given target position to the center of the end-effector that is reached by setting the joints according to the produced angle solution. The computation time is the time to calculate the angle solution given the target position. On the other hand, in the latter part (from Section~\ref{various_solutions} to Section~\ref{extended_applications}), we dive deeper into the proposed approach and quantify the characteristics of our method. Most experiments will be conducted using SIK and PSIK. However, if SIK and PSIK produce similar results, we pick only one to show. All experiments were run on a computer with an Intel i7-8700 CPU and 64 GB of memory.

\subsection{Comparison of distance error} \label{accuracy_performance}

We evaluate the performance of the methods by estimating the distance error of the end-effector over the entire working space. This is done by sampling 100 desired poses uniformly from the working space. The average distance error is then computed. For SIK, the average distance error is found to be about 0.5 cm, while for PSIK, the error is 1.7 cm. As mentioned in Section~\ref{ExpSetup}, PSIK is more difficult to train. The KL divergence prevented PSIK from fitting the training dataset, which is similar to a regularization term in training, so the accuracy of PSIK is limited. In contrast to our models, the FCN can only reach a 45 cm distance error, which is not surprised, because the various solutions in the same position stop the network from finding an effective association. However, even if we only keep one solution in each position, the FCN still suffers from poor performance, which only decreases about 5 cm distance error. We speculate that one possible reason is that the adjacent target positions may keep different postures so that the network may receive similar inputs, but their outputs are extremely different, and the weights are still in competition during training. The error of the ANFIS model is 70 cm, which is the worst performance among these methods.

We note that the dataset has different densities in different regions of the working space due to the nature of the robotic arm, i.e., based on the angles of the joints. Thus, to further examine the performance of our approach in different regions of the working space, we divided the working space with different sizes of circles centered around the base of the robot and sampled 100 random target positions on each of the circles. The resultant average distance errors are shown in Table~\ref{radius-table-SIK}. 

Table~\ref{radius-table-SIK} shows that SIK has worse performance when the end-effector is near the body of the arm. It performs even worse than PSIK. The best region for SIK is around 20 cm to 50 cm, where SIK has a 0.57 cm average distance error. Compared with SIK, the average distance errors in PSIK are relatively uniform across different regions because PSIK treats the dataset as distributions during training and considers nearby data points together. Both FCN models have higher performance when the radius is under 40 cm, which can also be seen in the results of our models. 

\begin{table*}
\caption{Comparison of the distance error of the end-effector with different radius}
\label{radius-table-SIK}
\centering
\resizebox{\textwidth}{!}{%
\begin{tabular}{llllllllllllllllllll} 
\hline
&Radius (cm)        & 5   & 10  & 15  & 20  & 25  & 30  & 35  & 40  & 45  & 50  & 55  & 60 & 65 & 70 & 75 & 80 & 85 & 90    \\
SIK (original dataset)&Error (cm)         & 1.84 & 1.87 & 0.42 & 0.44 & 0.73 & 0.47 & 0.71 & 0.63 & 0.58 & 0.68 & 0.64 & 0.64 & 0.84 & 0.65 & 0.73 & 0.83 & 1.15 & 0.88   \\ 
\hline
Average Distance Error (cm) & \multicolumn{18}{c}{\textbf{0.82}}  \\
\midrule
PSIK (original dataset)&Error (cm)         & 1.31 & 2.16 & 1.22 & 1.21 & 2.36 & 1.15 & 1.35 & 1.34 & 1.31 & 1.78 & 1.90 & 1.61 & 1.6 & 2.18 & 2.16 & 1.86 & 2.45 & 2.11   \\ 
\hline
Average Distance Error (cm) & \multicolumn{18}{c}{\textbf{1.73}}  \\
\midrule
ANFIS (uni-position dataset)&Error (cm)         & 91.1 & 88.6 & 88.2 & 85.8 & 82.2 & 82.5 & 75.9 & 76.5 & 74.7 & 70.1 & 70.3 & 71.3 & 68.3 & 66.2 & 66.5 & 62.9 & 60.5 & 68.4   \\ 
\hline
Average Distance Error (cm) & \multicolumn{18}{c}{\textbf{75.0}}  \\
\midrule
FCN (original dataset)&Error (cm)         & 26.2 & 23.9 & 28.6 & 27.4 & 29.0 & 29.9 & 30.5 & 37.1 & 40.6 & 35.3 & 41.8 & 50.9 & 59.5 & 56.4 & 66.2 & 70.0 & 76.3 & 80   \\ 
\hline
Average Distance Error (cm) & \multicolumn{18}{c}{\textbf{45.0}}  \\
\midrule
FCN (uni-position dataset)&Error (cm)         & 18.6 & 18.8 & 18.6 & 19.2 & 22.2 & 22.8 & 26.4 & 32.1 & 31.3 & 40.1 & 40.6 & 41.7 & 57.9 & 59.6 & 67.9 & 66.5 & 76.9 & 80.5   \\ 
\hline
Average Distance Error (cm) & \multicolumn{18}{c}{\textbf{41.2}}  \\
\hline
\end{tabular}%
}
\end{table*}

In this subsection, we find that the traditional learning-based methods easily suffer from poor performance when the dataset is unordered. This can explain why most previous studies usually collect a specialized training dataset, e.g., a continuous trajectory, to train their networks. By contrast, our approach exploits self-supervised learning to learn an extra index to solve the problem and free the wights from dilemmas during training.

\subsection{Computing Time} \label{time_performance}

To evaluate the computation time, we measure the execution time of running the SIK/PSIK models to generate the angle solutions. The Python \textit{time.clock()} function, which returns the current processor time in seconds, is used to estimate the computation time. We randomly sampled 100 testing points across the entire working space to estimate the average computation time. The results show that the SIK/PSIK models take about 0.004 seconds to obtain one solution. The fully-connected network takes about 0.003 seconds to obtain the angle solution. This is because the computations of learning-based methods are related to the number of wights. Although the lengths of the hidden layers are the same in both methods, the weights of our approach are slightly higher than the fully-connected network because the input layer of the decoder needs extra room for the posture indices. The ANFIS, on the other hand, only takes 0.00015 seconds to get the solution due to the small number of weights.

\subsection{Distance error in different postures} \label{various_solutions}

\begin{figure}
\centering
  \includegraphics[width=\linewidth]{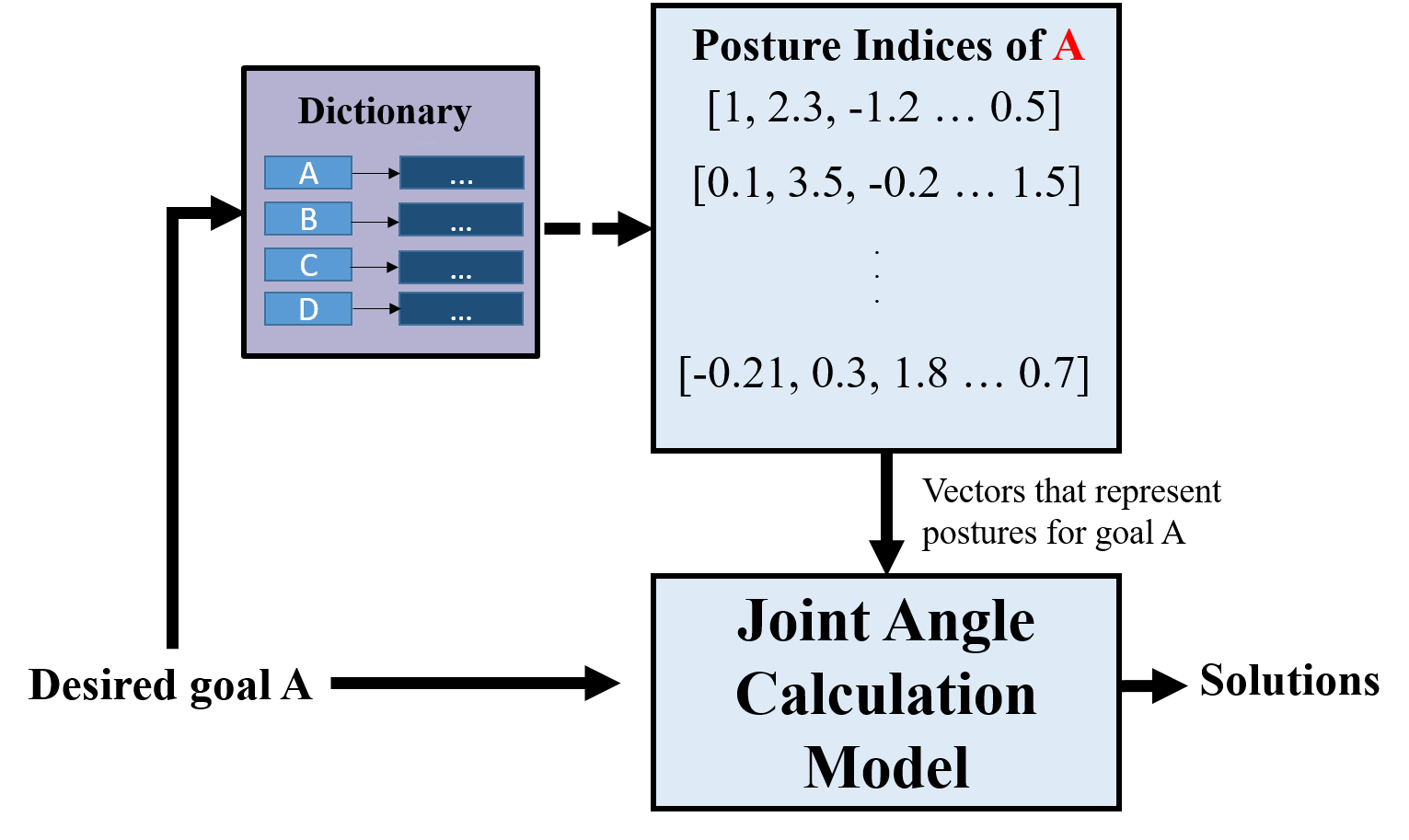}  
  \caption{The KD-tree and the dictionary to track positions in the dataset and their posture indices.}
  \label{sample_different_indices}
\end{figure}

Unlike traditional learning-based methods, the proposed method can obtain various angle solutions in the same target position. In this subsection, we evaluate the various angle solutions generated by the proposed methods. We want to investigate whether the proposed method can generate different poses and keep the same performance in all the solutions. This is done by randomly sampling one target position from each of the four quadrants of the working space and selecting ten different posture indices in each position. The problem here is that we do not know what values of indices can lead to quite different solutions. If we slightly adjust the posture indices, we may only obtain similar postures. If we arbitrarily give values to indices, the decoder may not generate high-quality solutions due to irrational postures. To obtain various robot postures instead of testing in similar postures, we use the KD-tree and the dictionary data structure to track the available indices in each pose in the training dataset, so that we can take the indices that are far away from each other to both ensure the diversity and rational  assignments (as shown in Fig.~\ref{sample_different_indices}).

We first evaluate the diversity of the angle solutions that we chose. Table~\ref{variance-table-SIK} shows the variance of the resultant angle solutions for each joint using SIK, PSIK, respectively. As shown in the table, Joint 1 has a high variance in every position, which means Joint 1 has quite different values in the ten angle solutions generated by the ten posture indices in all four positions. Since Joint 1 is nearest to the base of the arm, this implies that the robot may start by facing very different directions and eventually its end-effector can reach the same target position. If we examine Joints 1, 2, and 3 together, which determine the location of the elbow (Joint 4), we can also see that they have high variances. It means that Joint 4 can have very different positions while still taking the end-effector to the same target position. With respect to the diversity of the angles, SIK and PSIK have similar diversity in the four target positions.

\begin{table*}
\caption{The variance of the angle for each joint in the four target positions using SIK, and PSIK based on 10 different angle solutions. Joint 1 controls the base of the robotic arm, whereas Joint 7 controls the rotation of the end-effector (gripper).}
\label{variance-table-SIK}
\centering
\begin{tabular}{cllllllll} 
\toprule
&desired pose                     & ~ Joint 1~~ & ~ Joint 2~~ & ~ Joint 3~~ & ~ Joint 4~~ & ~ Joint 5~~ & ~ Joint 6~~ & ~ Joint 7~~  \\ 
\midrule
\multirow{4}{4em}{SIK}&~ ~ ~ ~ ~\textbf{a}     & 141.23      & 70.59      & 175.55     & 5.75        & 178.47      & 54.69       & 133.93         \\
&~ ~ ~ ~\textbf{~b}~~   & 194.92      & 56.36       & 188.62      & 15.60        & 213.78      & 147.51     & 114.54        \\
&~ ~ ~ ~ ~\textit{\textbf{c~ ~~}} & 166.66      & 99.60     & 176.19      & 2.99       & 142.76      & 109.93     & 51.89         \\
&~ ~ ~ ~ ~\textit{\textbf{d~ ~~}} & 110.25      & 57.55      & 255.89      & 1.25       & 335.27      & 131.81      & 126.34         \\
\midrule
\multirow{4}{4em}{PSIK}&~ ~ ~ ~ ~\textbf{a}     & 138.53      & 70.26       & 176.80      & 5.69        & 179.89      & 54.97       & 0.01         \\
&~ ~ ~ ~\textbf{ ~b}~~   & 205.18      & 56.01       & 159.10      & 17.50        & 249.05      & 121.48      & 71.94        \\
&~ ~ ~ ~ ~\textit{\textbf{c~ ~~}} & 164.97      & 98.49      & 178.18      & 3.03        & 142.63      & 109.57      & 0.01         \\
&~ ~ ~ ~ ~\textit{\textbf{d~ ~~}} & 172.87      & 55.37       & 249.41       & 1.28        & 341.17      & 131.19      & 1.50         \\
\toprule
\end{tabular}
\end{table*}

Table~\ref{visual-table} illustrates the ten postures generated from the ten posture indices to reach each of the four target positions using PSIK. From the table we can see that the robotic arm can approach the target position (red dot) from different directions with different heights of the elbow joint and even different orientations of the end-effector. The postures are quite distinct if examined visually. The postures produced by SIK exhibit similar behaviors.

\begin{table*}
\caption{Illustrations of the ten different angle solutions in the four target positions}
\label{visual-table}
\centering
\begin{tabular}{lllllllllll} 
\toprule
  & No. 1 & No. 2 & No. 3 & No. 4 & No. 5 & No. 6 & No. 7 & No. 8 & No. 9 & No. 10  \\ 
\midrule
\textbf{a} & \includegraphics[height=1.5cm, width=1cm]{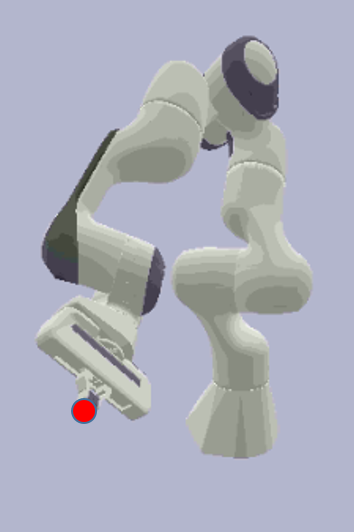}     & \includegraphics[height=1.5cm, width=1cm]{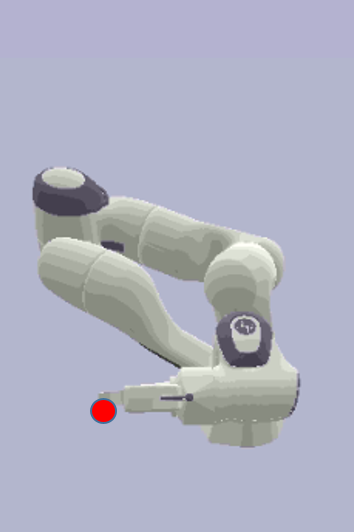}     & \includegraphics[height=1.5cm, width=1cm]{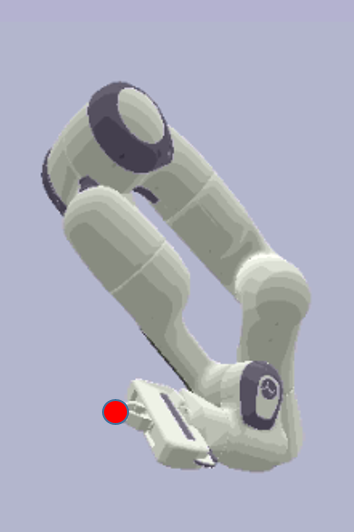}     & \includegraphics[height=1.5cm, width=1cm]{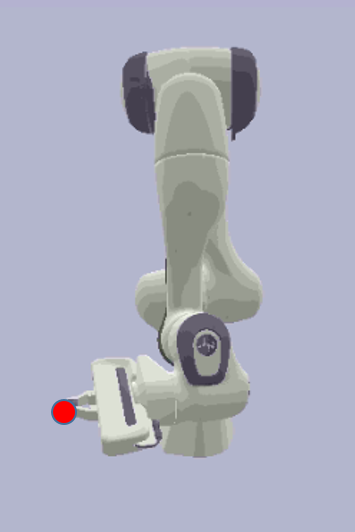}     & \includegraphics[height=1.5cm, width=1cm]{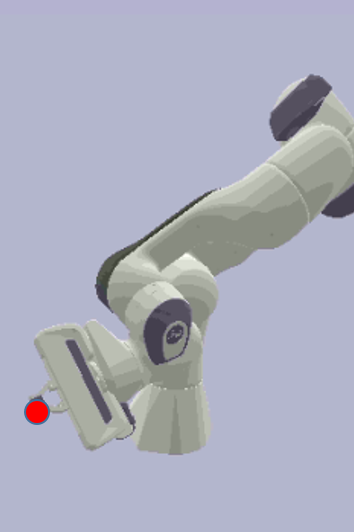}     & \includegraphics[height=1.5cm, width=1cm]{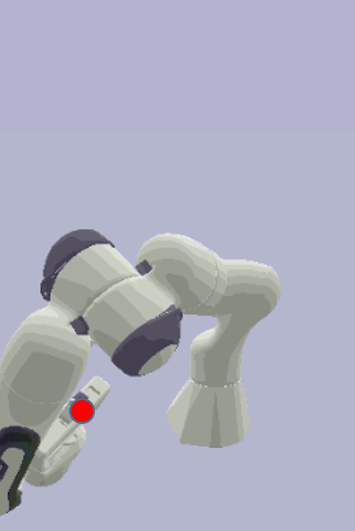}     & \includegraphics[height=1.5cm, width=1cm]{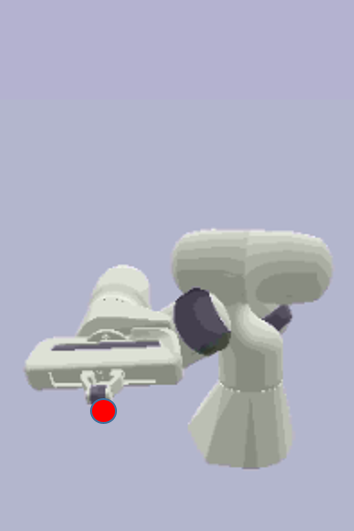}     & \includegraphics[height=1.5cm, width=1cm]{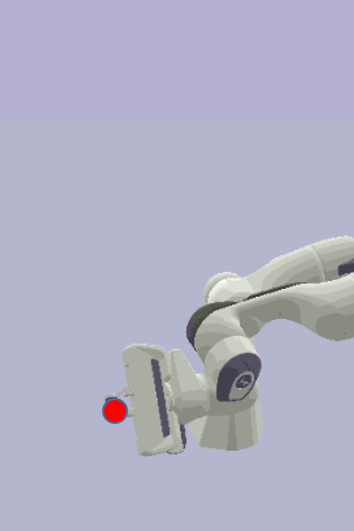}     & \includegraphics[height=1.5cm, width=1cm]{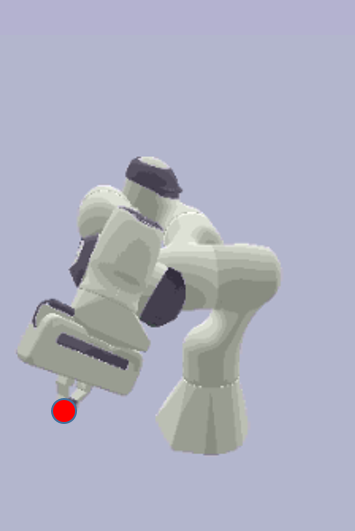}     & \includegraphics[height=1.5cm, width=1cm]{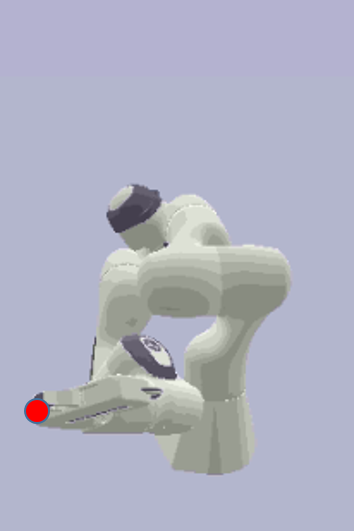}      \\
\textbf{b} & \includegraphics[height=1.5cm, width=1cm]{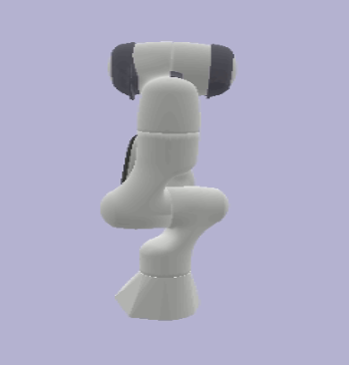}     & \includegraphics[height=1.5cm, width=1cm]{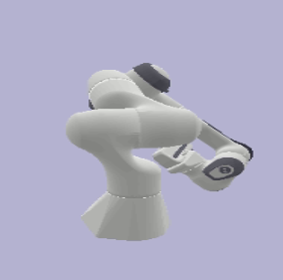}     & \includegraphics[height=1.5cm, width=1cm]{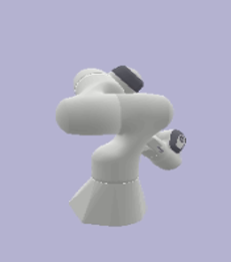}     & \includegraphics[height=1.5cm, width=1cm]{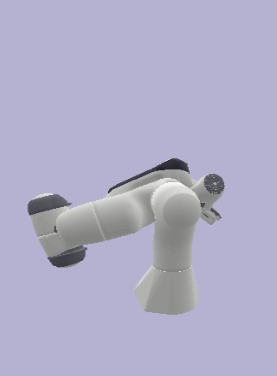}     & \includegraphics[height=1.5cm, width=1cm]{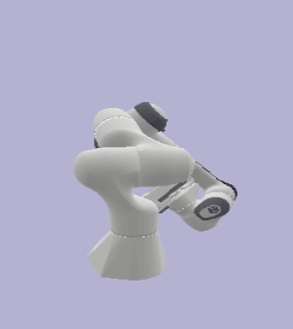}     & \includegraphics[height=1.5cm, width=1cm]{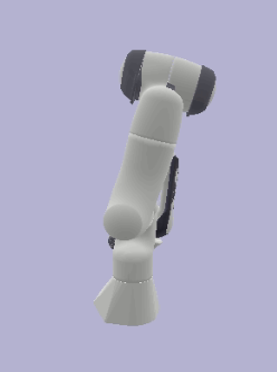}     & \includegraphics[height=1.5cm, width=1cm]{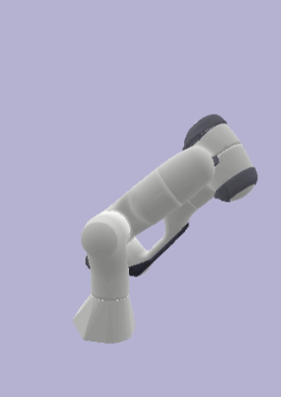}     & \includegraphics[height=1.5cm, width=1cm]{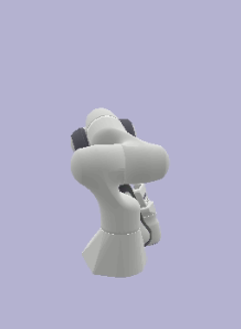}     & \includegraphics[height=1.5cm, width=1cm]{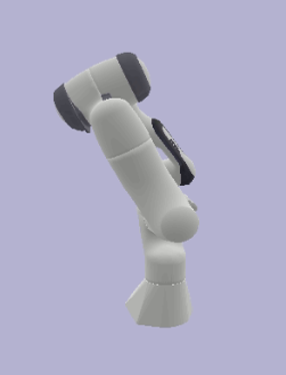}     & \includegraphics[height=1.5cm, width=1cm]{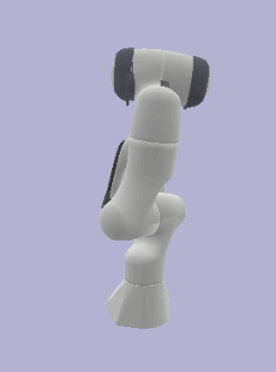}      \\
\textbf{c} & \includegraphics[height=1.5cm, width=1cm]{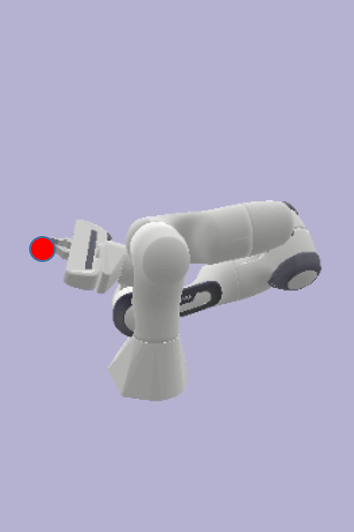}     & \includegraphics[height=1.5cm, width=1cm]{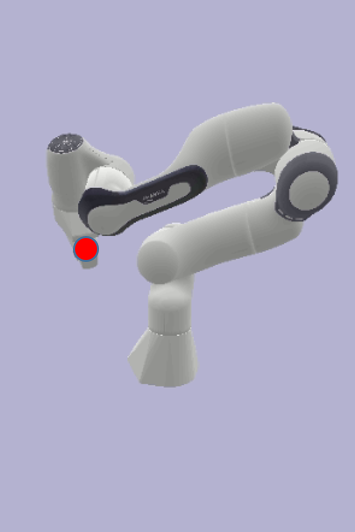}     & \includegraphics[height=1.5cm, width=1cm]{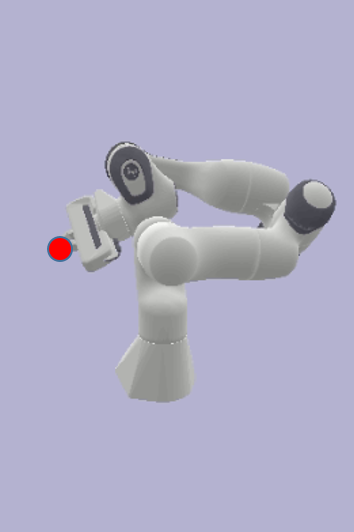}     & \includegraphics[height=1.5cm, width=1cm]{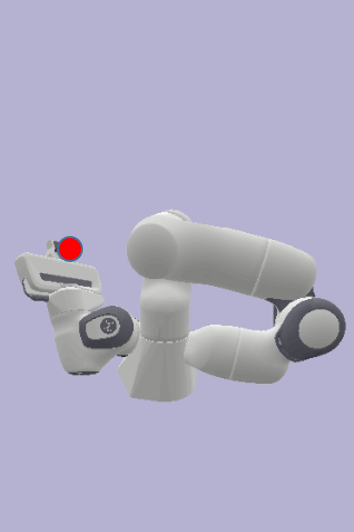}     & \includegraphics[height=1.5cm, width=1cm]{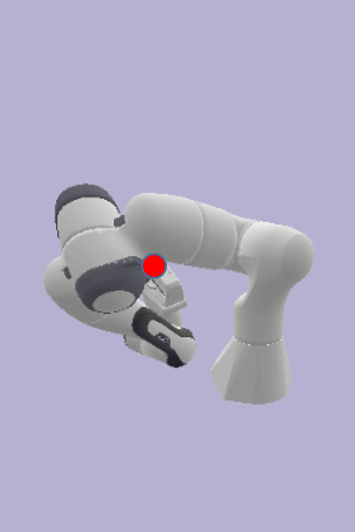}     & \includegraphics[height=1.5cm, width=1cm]{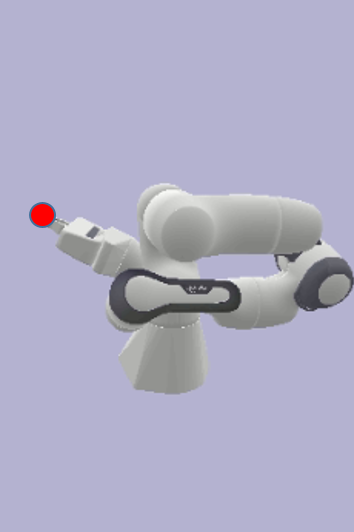}     & \includegraphics[height=1.5cm, width=1cm]{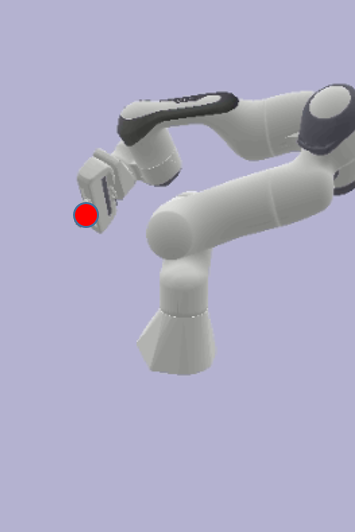}     & \includegraphics[height=1.5cm, width=1cm]{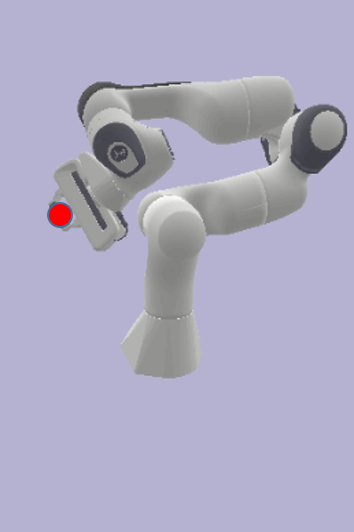}     & \includegraphics[height=1.5cm, width=1cm]{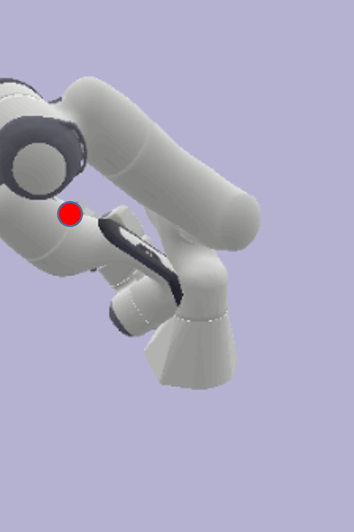}     & \includegraphics[height=1.5cm, width=1cm]{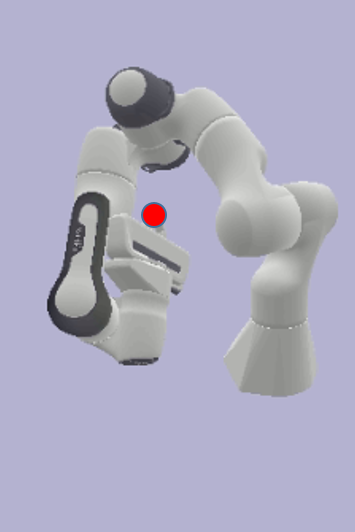}      \\
\textbf{d} & \includegraphics[height=1.5cm, width=1cm]{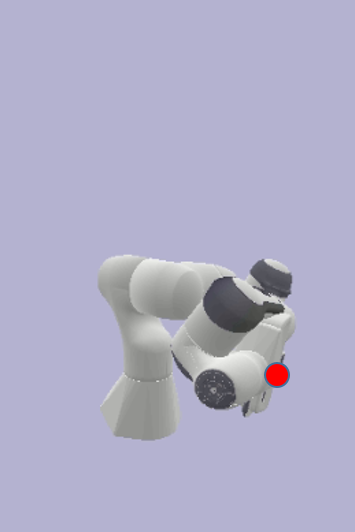}     & \includegraphics[height=1.5cm, width=1cm]{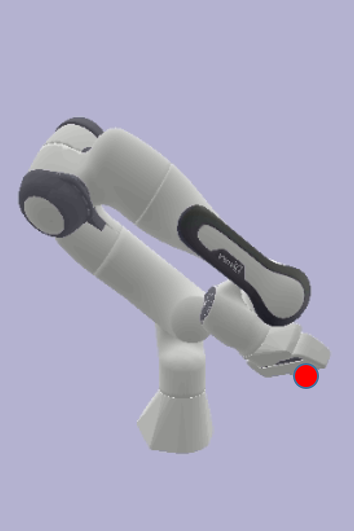}     & \includegraphics[height=1.5cm, width=1cm]{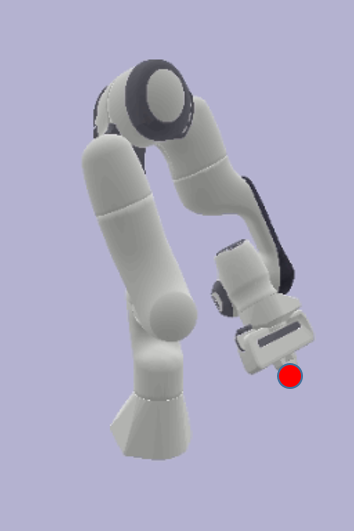}     & \includegraphics[height=1.5cm, width=1cm]{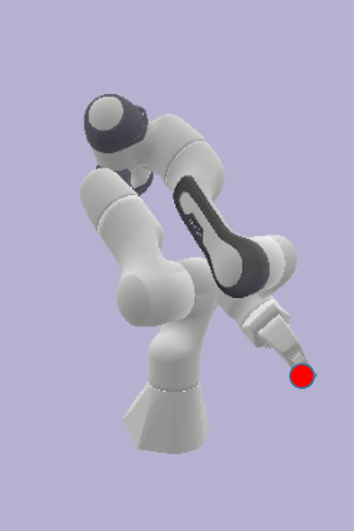}     & \includegraphics[height=1.5cm, width=1cm]{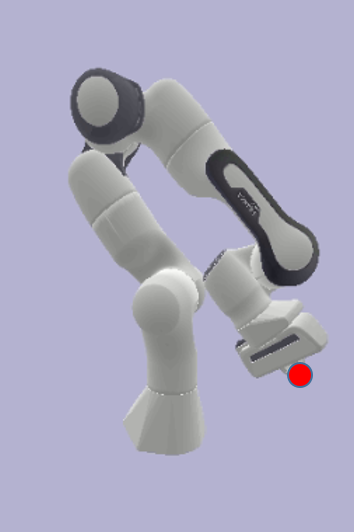}     & \includegraphics[height=1.5cm, width=1cm]{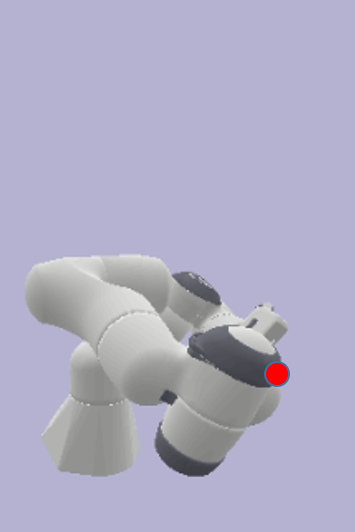}     & \includegraphics[height=1.5cm, width=1cm]{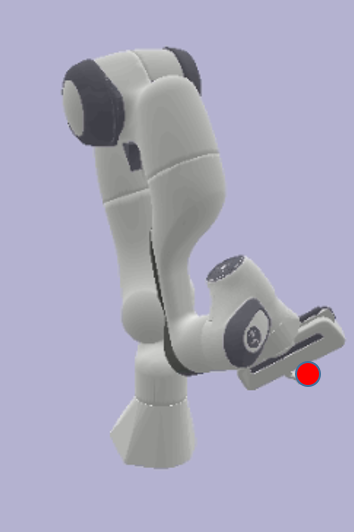}     & \includegraphics[height=1.5cm, width=1cm]{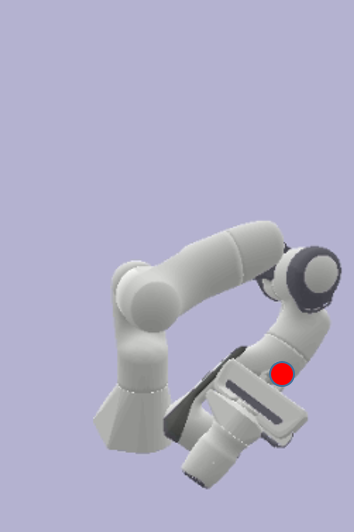}     & \includegraphics[height=1.5cm, width=1cm]{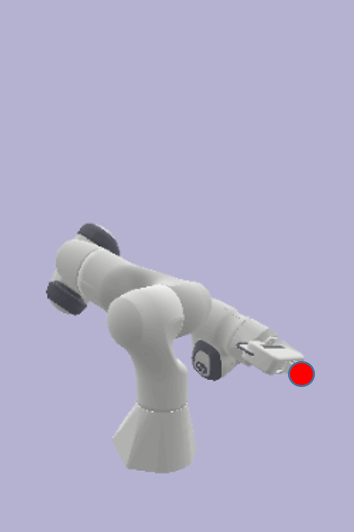}     & \includegraphics[height=1.5cm, width=1cm]{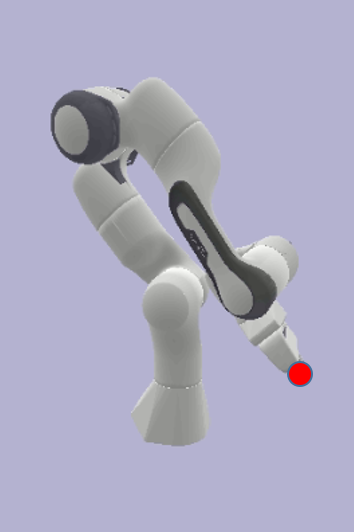}      \\
\bottomrule
\end{tabular}
\end{table*}

Table~\ref{error-table-SIK} shows more details of the four target positions and their distance errors using SIK and PSIK. The first column is the positions in meters, and the remaining columns show the Euclidean distance errors of the end-effector in centimeters. In SIK, the average distance errors are between 0.4 cm to 0.5 cm, whereas PSIK has average distance errors between 1.1 cm to 1.5 cm.

To sum up, the proposed methods can model the redundant robotic arm and keep multiple angle solutions to reach a given target position, and the average distance errors of the end-effector are about 0.5 cm and 1 cm. SIK and PSIK can both produce distinct, controllable, and reproducible postures.

\begin{table*}
\caption{The distance errors (cm) of the ten angle solutions in the four target positions using SIK and PSIK}
\label{error-table-SIK}
\centering
\setlength\tabcolsep{2pt}
\begin{tabular}{lllllllllllll} 
\toprule
&Goal (meter):~ (x, y, z)~ & No. 1 & No. 2 & No. 3 & ~No. 4 & No. 5 & No. 6 & No. 7 & No. 8 & No. 9 & No. 10 &  \textbf{Avg.}  \\ 
\midrule
\multirow{4}{4em}{SIK}&(0.1357, -0.321, 0.222)   & 0.22  & 0.25  & 0.54  & 0.22   & 0.31  & 1.6  & 0.29  & 0.56  & 0.19  & 0.61   & 0.48              \\
&(-0.333, 0.246, 0.1999)   & 0.35  & 0.25  & 0.48  & 0.4   & 0.36  & 0.38  & 0.42  & 0.4  & 0.34  & 0.35   & 0.37               \\
&(-0.188, -0.168, 0.321)   & 0.74  & 0.24  & 1.04  & 0.57   & 0.40  & 0.63  & 0.43  & 0.45  & 0.31  & 0.61   & 0.54               \\
&(0.2020, 0.2021, 0.2022)  & 0.21  & 0.29  & 0.31  & 0.47   & 1.08  & 0.30  & 0.46  & 0.40  & 0.46  & 0.71   & 0.47               \\

\midrule
\multirow{4}{4em}{PSIK}&(0.1357, -0.321, 0.222)   & 0.98  & 0.65  & 1.03  & 0.82   & 1.70  & 1.28  & 3.24  & 0.63  & 1.20  & 2.72   & 1.43              \\
&(-0.333, 0.246, 0.1999)   & 0.93  & 1.51  & 1.46  & 2.44   & 1.32  & 0.40  & 0.74  & 1.48  & 1.13  & 1.83   & 1.32               \\
&(-0.188, -0.168, 0.321)   & 0.94  & 1.45  & 3.35  & 1.26   & 0.51  & 1.64  & 0.77  & 1.09  & 0.29  & 0.63   & 1.19               \\
&(0.2020, 0.2021, 0.2022)  & 1.03  & 0.27  & 1.04  & 2.18   & 2.05  & 1.31  & 1.18  & 1.39  & 2.94  & 1.17   & 1.46               \\
\cmidrule[\heavyrulewidth]{1-7}\cmidrule[\heavyrulewidth]{8-11}\cmidrule[\heavyrulewidth]{12-13}
\end{tabular}
\end{table*}

\subsection{Impacts of Posture Indices} \label{section_posture_indices}

Our approach represents the postures of the robotic arm numerically using the 4-dimensional posture indices. Intuitively, every element in the posture indices should affect some aspects of the final posture of the arm. Unfortunately, the unsupervised learning that we used to obtain the posture indices lacks semantic explanations, though it allows our approach to generalize to any DoF robotic arms. This section tries to exploit the properties of the posture indices and compares those produced by SIK and PSIK. 

The first question we attempt to answer is: "What dimensionality should the posture index have?" Since high-DoF IK is very new in robotics, we cannot find any related literature to assist us in answering this question. Hence, we tried different dimensionalities from one to seven in the experiments, because we used a 7-DoF robotic arm as the target robotic arm and it is unreasonable if the dimension of the posture indices is greater than the number of the joints. Our experiments show that the models with a dimension smaller than three failed to fit the training data. The loss values of these models stopped decreasing in an early stage. On the other hand, the models with a dimension greater than or equal to three can successfully achieve a good comparable performance. The main difference among them is the speed to converge. The longer the indices are, the faster the loss value decreases. Since the models with 3-dimensional indices need more hyper-parameter adjustments and training time, we therefore use 4-dimensional posture indices in most experiments discussed in this section. 

\begin{figure}
\centering
  \includegraphics[width=\linewidth]{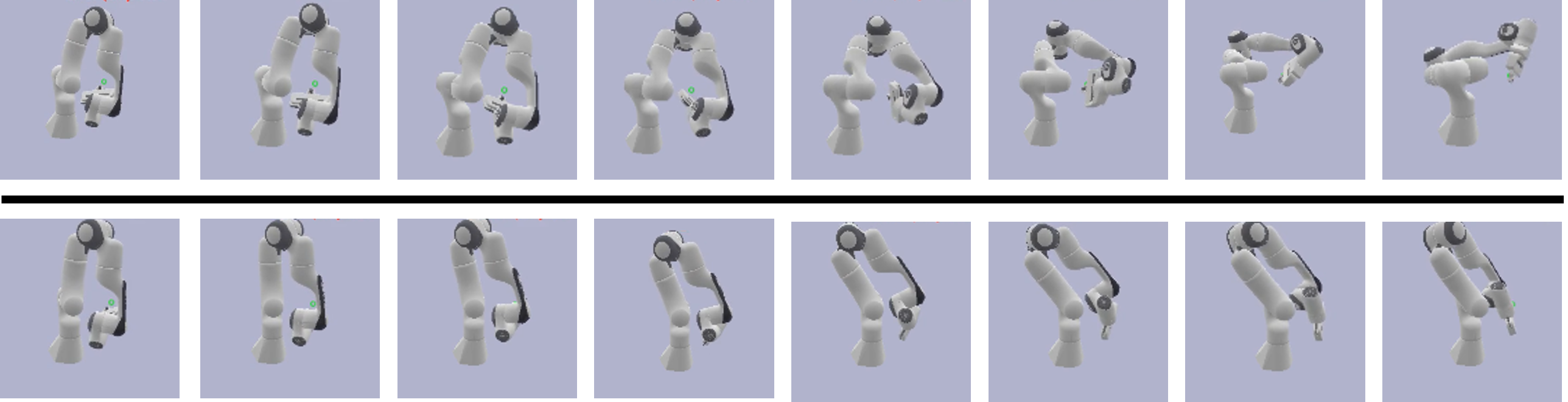}  
  \caption{Since we know which elements in the posture index control the orientation of the end-effector, we can adjust the elements to make the end-effector move along slant lines and still keep touching the target position.}
  \label{slant-lines}
\end{figure}

Next, we study the effects of changing different elements in the posture indices. Let us start with 4-dimensional posture indices. Table~\ref{second_posture_index_table} shows the changes of the postures when different elements in a posture index are changed for the target position (-0.25, -0.25, 0.25). There are two different trends in the postures when the robotic arm tries to reach the given target position. One affects the orientation of the end-effector (elements 1 \& 4), and the other affects the body of the robotic arm but keeps both the orientation and the location fixed (elements 2 \& 3). More specifically, element 1 makes the end-effector rotate around the vertical line, whereas element 4 rotates it around the horizontal line. In other words, we can change the two elements to control the direction that the end-effector approaches the target position. Fig.~\ref{slant-lines} shows the results when we adjust both elements 1 and 4. As we can see, the end-effector moves along slant lines. 

On the other hand, elements 2 and 3 of the posture indices change the slope of the robotic arm. They cause similar effects except for the side. Such effects are a result of the extra redundancy in high-DoF robotic arms. Note that robotic arms with DoF fewer than seven can only have one solution to reach a target position when they need to keep a fixed orientation.

\begin{table*}
\caption{Attributes of the posture indices in target position (-0.25, -0.25, 0.25)}
\label{second_posture_index_table}
\centering
\begin{tabular}{lllllll} 
\toprule
 Element 1& \includegraphics[height=1.5cm, width=1.5cm]{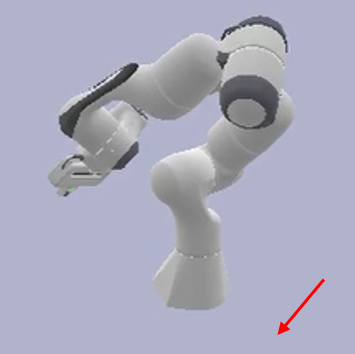} & \includegraphics[height=1.5cm, width=1.5cm]{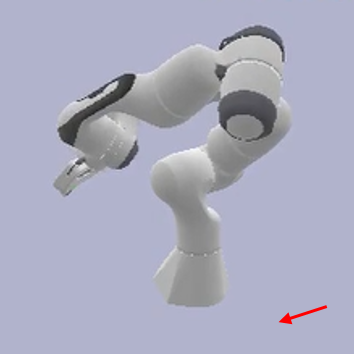}  & \includegraphics[height=1.5cm, width=1.5cm]{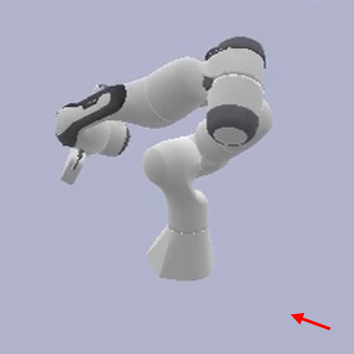} &\includegraphics[height=1.5cm, width=1.5cm]{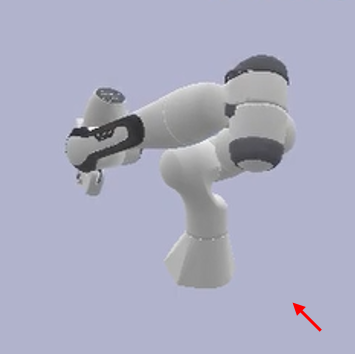}  & \includegraphics[height=1.5cm, width=1.5cm]{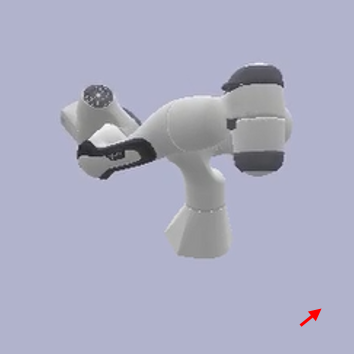} & \includegraphics[height=1.5cm, width=1.5cm]{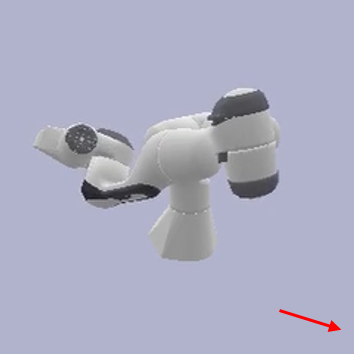}  \\
 Error (cm) & 0.08 & 0.24& 0.6& 0.81& 0.65& 0.51  \\ 
\midrule
 Element 2& \includegraphics[height=1.5cm, width=1.5cm]{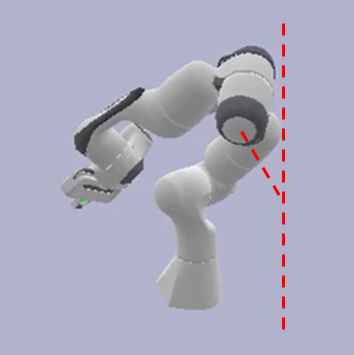} & \includegraphics[height=1.5cm, width=1.5cm]{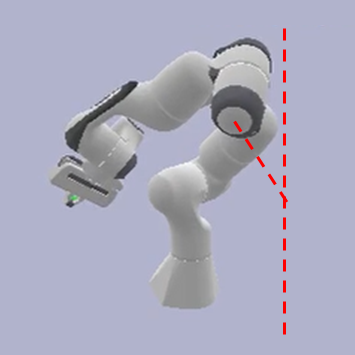} & \includegraphics[height=1.5cm, width=1.5cm]{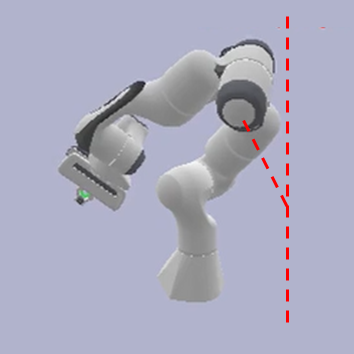} & \includegraphics[height=1.5cm, width=1.5cm]{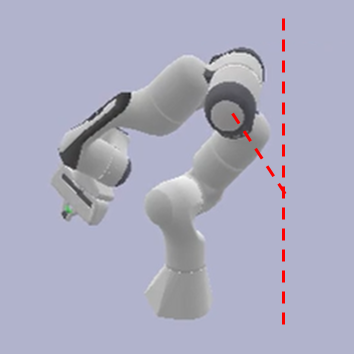} & \includegraphics[height=1.5cm, width=1.5cm]{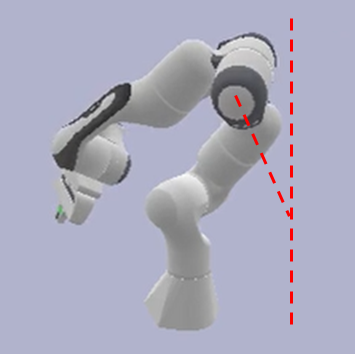} & \includegraphics[height=1.5cm, width=1.5cm]{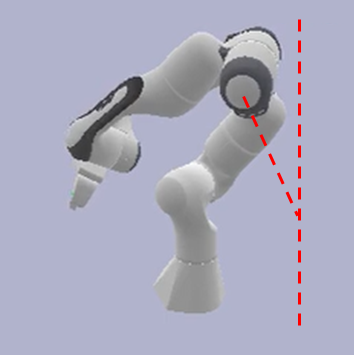}  \\
 Error (cm)& 0.06& 0.15 &0.18& 0.08& 0.49& 0.6  \\ 
\midrule
 Element 3& \includegraphics[height=1.5cm, width=1.5cm]{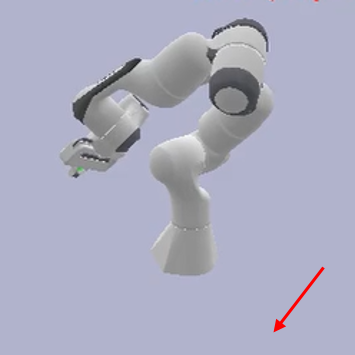} & \includegraphics[height=1.5cm, width=1.5cm]{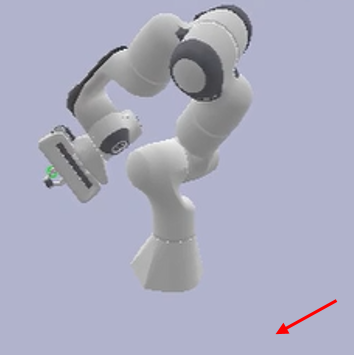} &\includegraphics[height=1.5cm, width=1.5cm]{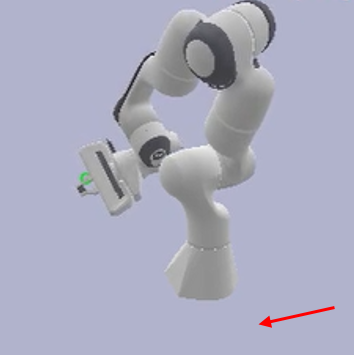}  & \includegraphics[height=1.5cm, width=1.5cm]{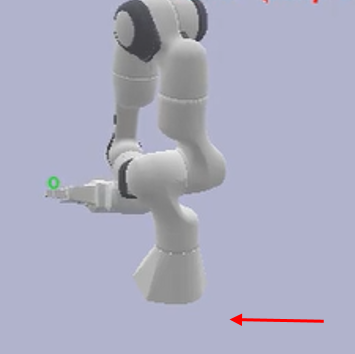} &\includegraphics[height=1.5cm, width=1.5cm]{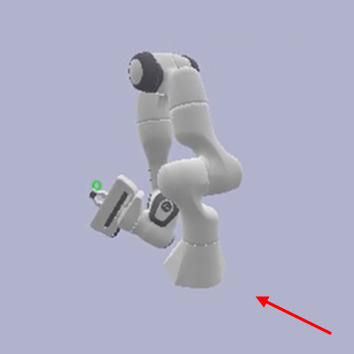}  & \includegraphics[height=1.5cm, width=1.5cm]{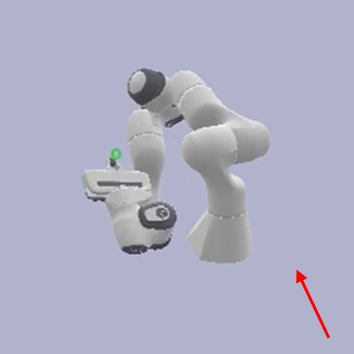}  \\
 Error (cm)& 0.03& 0.58& 0.66& 0.38& 0.37 &0.38  \\ 
\midrule
 Element 4& \includegraphics[height=1.5cm, width=1.5cm]{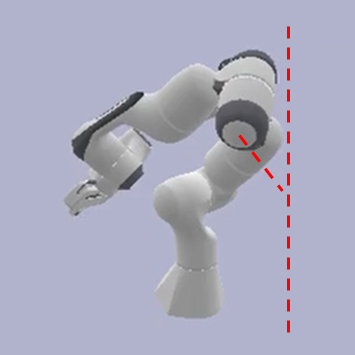} &  \includegraphics[height=1.5cm, width=1.5cm]{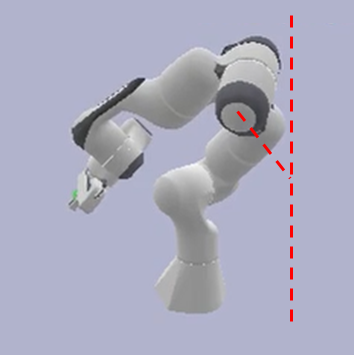} &  \includegraphics[height=1.5cm, width=1.5cm]{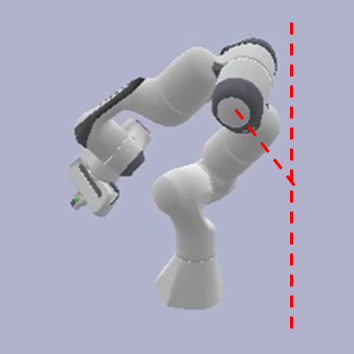} &  \includegraphics[height=1.5cm, width=1.5cm]{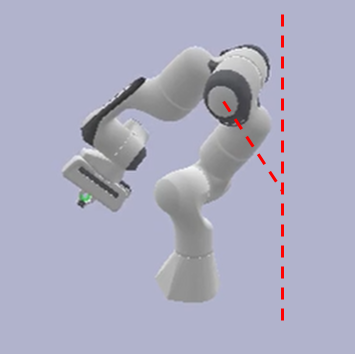} &  \includegraphics[height=1.5cm, width=1.5cm]{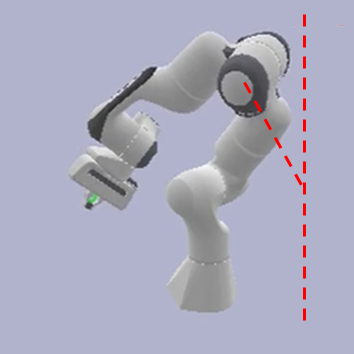} &  \includegraphics[height=1.5cm, width=1.5cm]{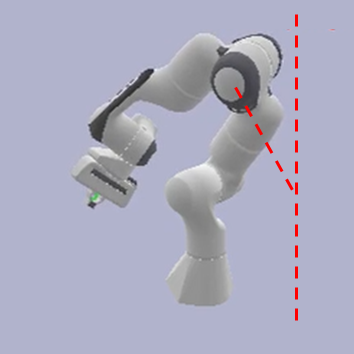}   \\
 Error (cm)& 0.03 & 0.11 & 0.18 & 0.19 & 0.42 & 0.86  \\
\bottomrule
\end{tabular}
\end{table*}

We next study how the roles of the elements in the posture indices change when the dimensionality of the indices is changed. We examine 3-dimensional posture indices first and observe the effects of adjusting each of the three elements. It is found that two of the three elements affect the orientation. This implies that they are essential so that the end-effector can approach the target position from any direction in a 3D space. The remaining element alters the slope, which means that the two elements that do the same in 4-dimensional posture indices are now merged into one due to their similarity. This also explains why the dimension of the posture indices has to be at least three. 

When the dimensionality of posture indices increases, the trained models can have more room to contain the two types of changing trends, which eases the difficulty of training. However, many elements in the posture indices may cause similar effects. In summary, the dimensionality of posture indices should be at least two (for affecting the orientation) plus the DoF of the target robotic arm minus six, i.e., the number of extra redundancies. 

Finally, we give a brief qualitative comparison of SIK and PSIK, after we have examined various aspects of these two methods. Both SIK and PSIK can obtain distinct postures for given poses. With respect to accuracy, SIK is much better than PSIK and is easier to train, because PSIK has more constraints and requires the posture indices to obey probabilistic distributions. However, PSIK has very uniform distance errors across the entire working space. This stabilizes the overall performance and extends the dataset smoothly to overcome the sparsity in collected training data. 

Regarding the characteristics of the obtained posture indices, SIK generates arbitrary floating-point numbers, whereas PSIK produces normalized numbers. If the application is just to obtain different angle solutions, then the normalization will not change the usage. We only need to retrieve the numbers from the dictionary and feed them into the joint angle calculation models to get the angle solutions. However, if the application requires changing the posture of the robotic arm from a given initial posture, then PSIK may be a better choice. The normalization of the posture indices means that all the numbers of the indices are bounded in a known range and the impacts of the postures are equally distributed in the range. Therefore, users have better controls in adjusting the elements in the posture indices.

\subsection{Including Orientation as Goal} \label{orientation_performance}

In this subsection, we examine SIK when the desired poses include target position as well as orientation. The implementation is the same as those presented in Section~\ref{ExpSetup} except for the length of the inputs to the joint angle calculation models, which becomes six plus the size of the posture indices. As mentioned in Section~\ref{section_posture_indices}, we found that the dimension of the posture indices has to be at least three, in which one element controls the slope of the arm and the other two control the orientation of the end-effector. Now, the goal requires orientation also, so the orientation of the end-effect is given. That means we cannot change the orientation for such goals. It follows that we may be able to reduce the length of the posture indices from three to one.

We evaluated the distance errors and the cosine similarity of the orientation of the end-effector in four different desired poses using 1-dimensional posture indices. The average distance error is 1.35 cm, and the average cosine similarity is 0.9983, which means the deviation of the orientation is about 3 degrees. This experiment validates that SIK can also be applied if the desired pose includes both target position and orientation. Table~\ref{vision-ori-table} illustrates the five postures generated from the five posture indices to reach the four target positions while maintaining the given orientations. We can easily see the similar effects as discussed in the previous sections considering only target positions. We also trained SIK using 3- and 4-dimensional posture indices. It turned out that the extra elements cause the same changing trend, i.e., slope adjustment, but not orientation.

 \begin{table*}
\caption{Illustrations of the five different angle solutions in the four target positions considering both target position and orientation using 1-dimensional posture indices}
\label{vision-ori-table}
\centering
\begin{tabular}{llllll} 
\toprule
  & No. 1 & No. 2 & No. 3 & No. 4 & No. 5  \\ 
\midrule
\textbf{a} & \includegraphics[height=1.5cm, width=1.5cm]{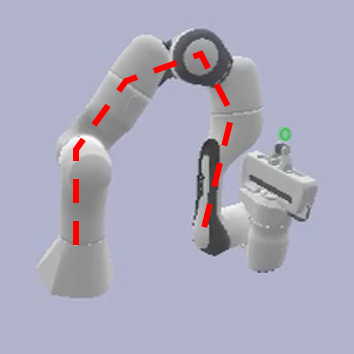}     & \includegraphics[height=1.5cm, width=1.5cm]{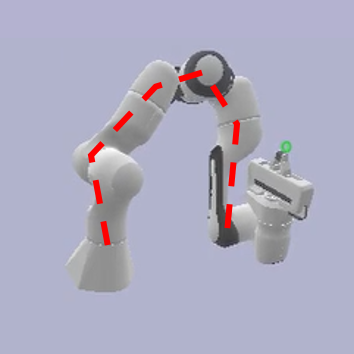}     & \includegraphics[height=1.5cm, width=1.5cm]{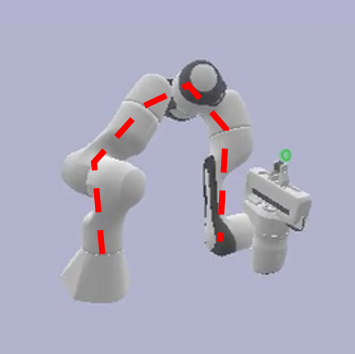}     &  \includegraphics[height=1.5cm, width=1.5cm]{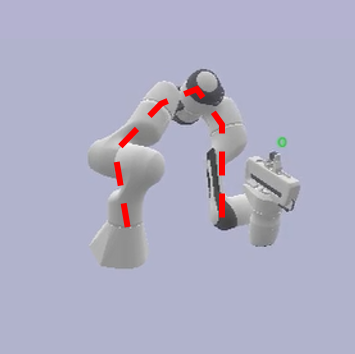}     & \includegraphics[height=1.5cm, width=1.5cm]{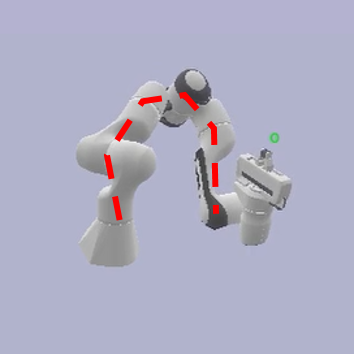}      \\
\textbf{b} & \includegraphics[height=1.5cm, width=1.5cm]{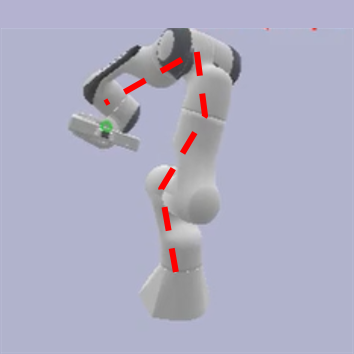}     & \includegraphics[height=1.5cm, width=1.5cm]{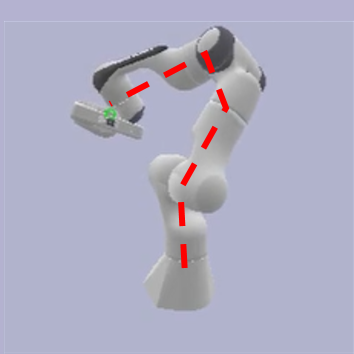}     & \includegraphics[height=1.5cm, width=1.5cm]{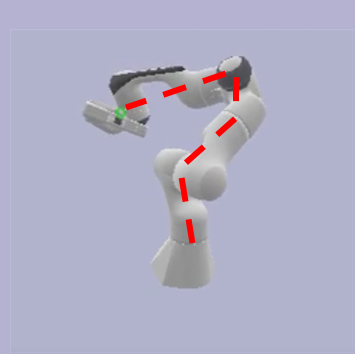}     &  \includegraphics[height=1.5cm, width=1.5cm]{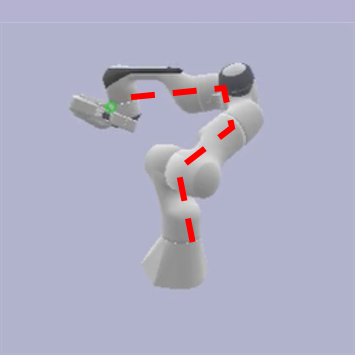}     & \includegraphics[height=1.5cm, width=1.5cm]{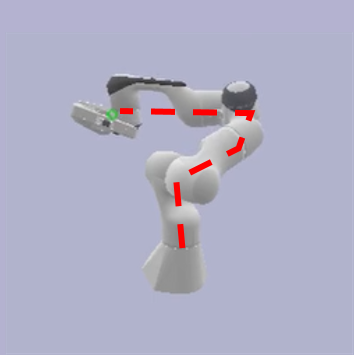}      \\
\textbf{c} & \includegraphics[height=1.5cm, width=1.5cm]{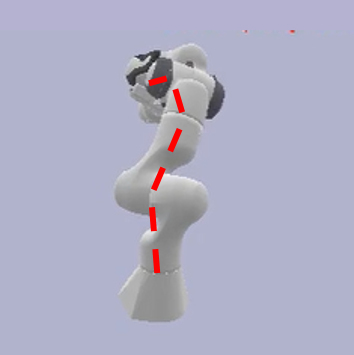}     & \includegraphics[height=1.5cm, width=1.5cm]{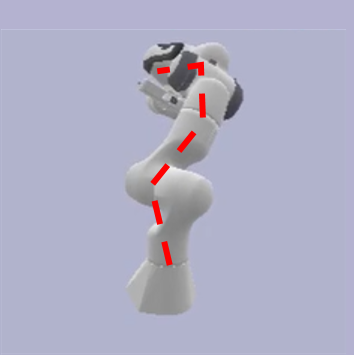}     & \includegraphics[height=1.5cm, width=1.5cm]{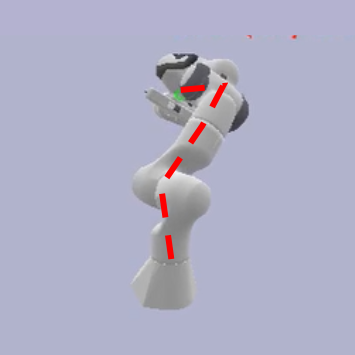}     &  \includegraphics[height=1.5cm, width=1.5cm]{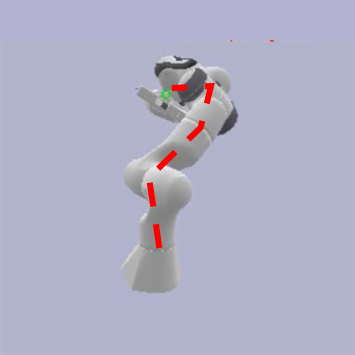}     & \includegraphics[height=1.5cm, width=1.5cm]{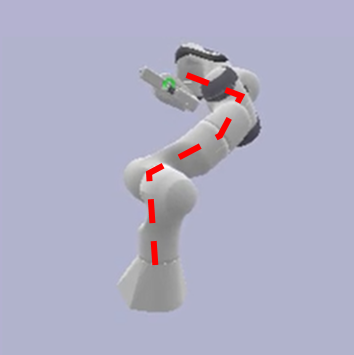}      \\
\textbf{d} & \includegraphics[height=1.5cm, width=1.5cm]{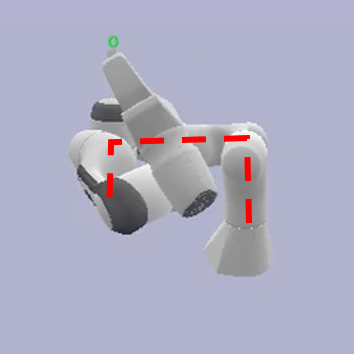}     & \includegraphics[height=1.5cm, width=1.5cm]{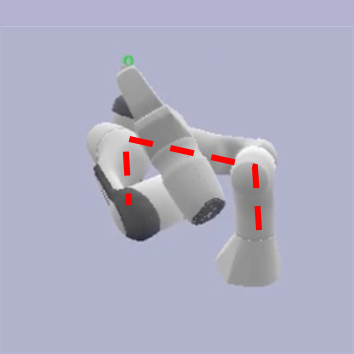}     & \includegraphics[height=1.5cm, width=1.5cm]{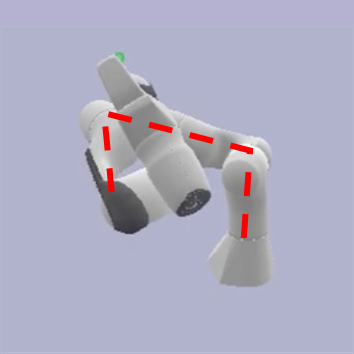}     &  \includegraphics[height=1.5cm, width=1.5cm]{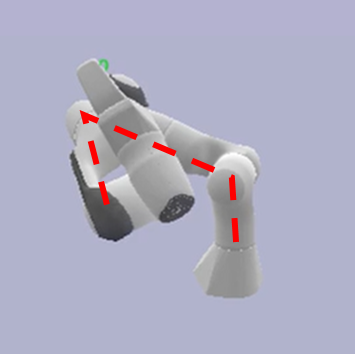}     & \includegraphics[height=1.5cm, width=1.5cm]{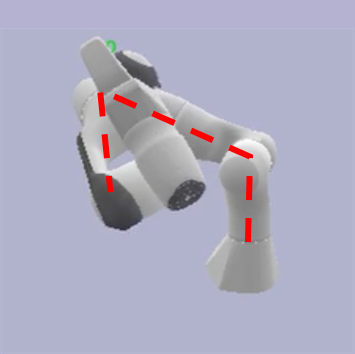}      \\
\bottomrule
\end{tabular}
\end{table*}

\subsection{Extended applications} \label{extended_applications}

\begin{figure}
  \center
  \includegraphics[width=\linewidth]{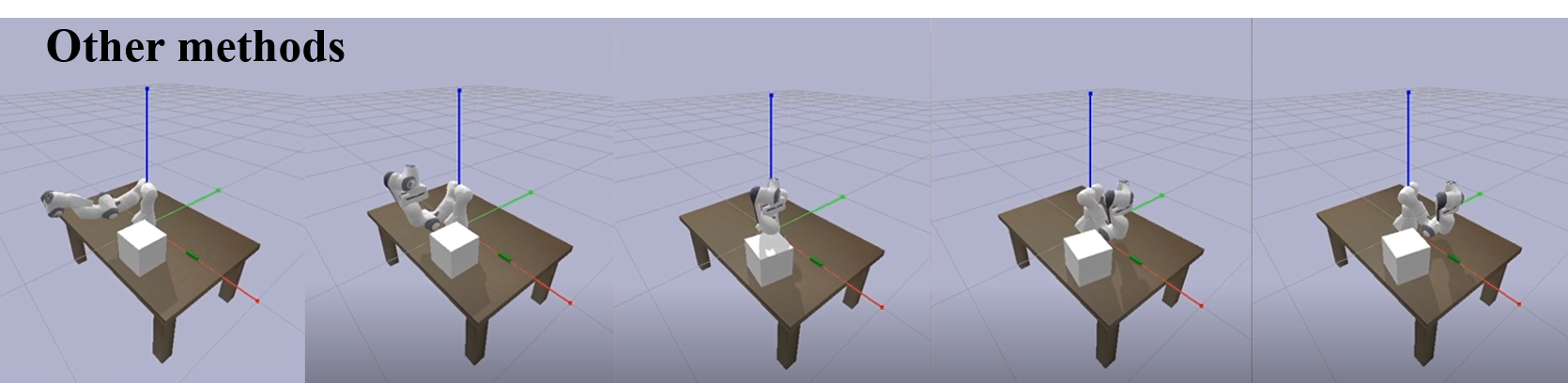}  
  \includegraphics[width=\linewidth]{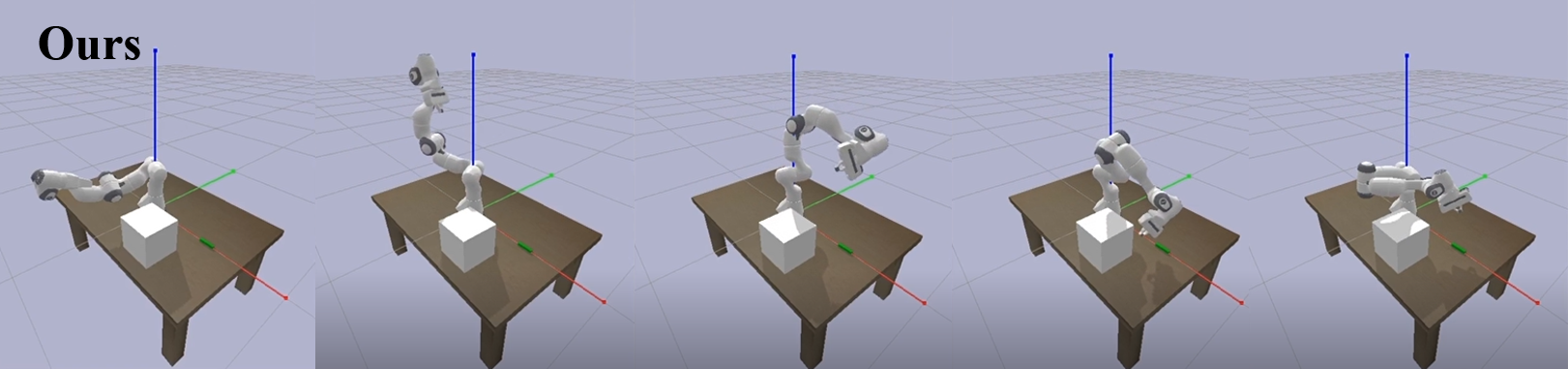}  
  \caption{An example of approaching the target position without colliding with the cube.}
  \label{real_example}
\end{figure}

In the last subsection, we show some concrete demonstrations of using other angle solutions to accomplish missions with extra constraints. Fig.~\ref{real_example} demonstrates an application of SIK for obstacle avoidance in a static environment. In this case study, we assume all the positions of obstacles are already known and fixed, and the task is to reach a given target position. In the example, there is a cube on the table, and the arm has to move the end-effector from the initial point (left) to the destination (right). Since the proposed SIK provides several different angle solutions to reach the target position, we can choose different solutions until we find one that does not run into the obstacle and execute it. The other two methods only have one way to reach the target, so it is hard for them to satisfy this kind of requirement. A common solution for them is to collect a specialized avoiding dataset after the environment is determined and retrain the models. In contrast to them, our approach only needs to train once and can be applied to different environments.

\section{Conclusions and Future Works}

We present in this paper a new data-driven approach to solving the IK problem for the entire working space of a robotic arm with redundant DoFs. The process to collect the training dataset can be fully automated with a few lines of code. Our approach can obtain multiple angle solutions for a given target position (plus a target orientation) and has a low distance error (${\sim}$0.5 cm) for any position in the entire working space of the robot. The proposed methods do not impose any special constraints on training data and thus can be applied to any DoF robotic arm. Compared with the latest data-driven methods, our work signifies a major step forward in solving the IK problem for robotic arms. This is the first data-driven research to address multiple angle solutions in IK, and a comprehensive study of the various properties of the proposed method is provided.

This work can find a diverse set of applications in robotic arms. If there are two different high-DoF robotic arms, it is possible to use the posture indices to make one arm imitate the other one. The two arms not only follow the same trajectories by the end-effector but also have similar postures. This could be very useful when one wants to transfer the skills from one robot to another. Another application of our work is that we can use the posture indices to adjust the robotic arm until it satisfies extra conditions in the environment. For example, if the environment contains obstacles and the elbow of the arm might collide with some obstacles during movement, then we can manipulate the posture indices to change the height of the elbow to avoid the collision. Another possibility is to choose the best posture index that can achieve certain optimization goals, e.g., reaching a goal with minimal power. 

Although this is the first paper to tackle the IK problem in the entire working space without designing customized datasets, and the accuracy is not our major concern, we still wonder whether we can further improve the accuracy of the model. This work uses a sparse training dataset (by every $ 30^\circ $ per joint) and has a subcentimeter distance error (${\sim}$0.5 cm). In future work, we will collect more data with shorter intervals to examine the performance of the proposed method. Furthermore, in this work, we do not propose an efficient way to adjust the posture indices for achieving tasks during inference. We will combine the properties of the posture indices with evaluation functions to control robotic arms more efficiently in the future.

\appendices

\section{Training with a Sequence of Integers as Indices}\label{simple_integer_index}

\begin{figure}
\centering
  \includegraphics[width=\linewidth]{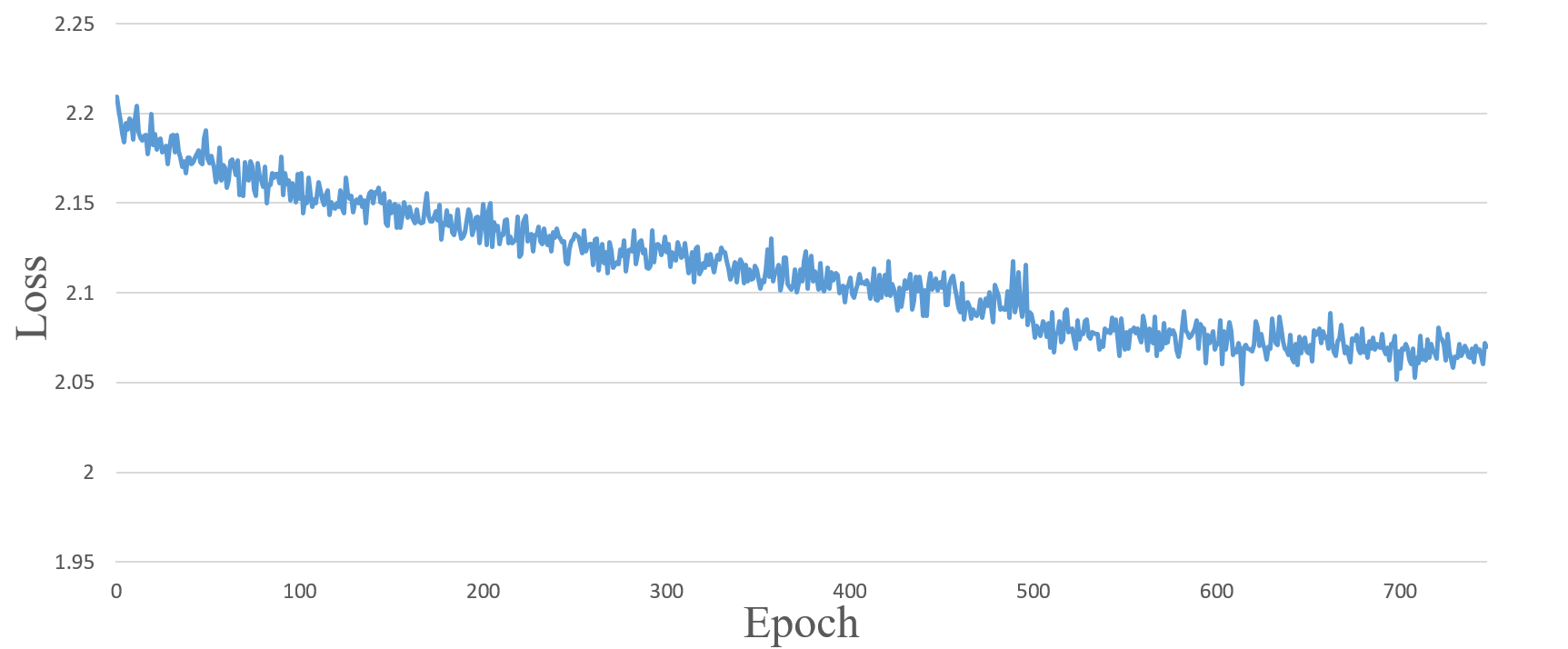}  
  \caption{Training loss with a sequence of integers as indices.}
  \label{training_loss}
\end{figure}

In addition to the proposed approach, in the early stage, we also attempted to accomplish this study through some intuitive manners, such as using integers as indices instead of learning them. Compared with the data collection proposed in Section~\ref{method_trainingdata}, we add an extra integer label before each data point. The integer starts with 1 to N for every data point in the same pose. During training, we attempted to directly train the joint angle calculation model with poses and their integer labels. The total number of training epochs is one thousand. As shown in Fig.~\ref{training_loss}, the model can not reduce the loss under 2 radians, which means the average reconstruction loss is more than 114 degrees. One possible reason is that the collected angle solutions can be considered random allocation, so the network cannot associate such randomness with the sequence of integers. In some cases, the posture pointed by an index may be quite different from the postures pointed by the neighboring indices. In others, similar indices may refer to similar postures.

\section{Training SIK With a Small Network} 

In the experimental section, we use a huge network to avoid considering if the network is too small to accommodate the data points in the entire working space. Here we implement SIK with a small network to see the degradation of our method. The encoder in the small network had  64, 64, 64, and 4 neurons in the five layers, whereas the decoder had 64, 64, 64, and 7 neutrons. The number of the weights of the network is one-millionth of the network in the experimental section. The training batch size was 64, and other parameters were not changed. The final average distance error over the entire working space is 19 cm, which is half of the FCN's error.

\bibliographystyle{plain}
\medskip
{
\small
\bibliography{ref}
}

\newpage

\begin{IEEEbiography}[{\includegraphics[width=1in,height=1.25in,clip,keepaspectratio]{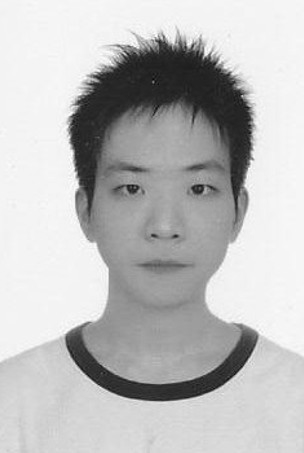}}]{Chi-Kai Ho} received the bachelors degree from the Department of Computer Science, National University of Kaohsiung , Taiwan. He is currently pursuing the Ph.D. degree with the Department of Computer Science, National Tsing Hua University, Taiwan. His current research interests include reinforcement learning and robotic manipulation.

\end{IEEEbiography}

\begin{IEEEbiography}[{\includegraphics[width=1in,height=1.25in,clip,keepaspectratio]{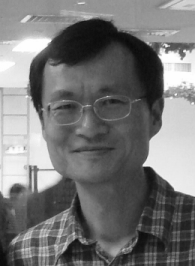}}]{Chung-Ta King} received the B.S. degree in electrical engineering from National Taiwan University, Taiwan, R.O.C., in 1980, and the M.S. and Ph.D. degrees in computer science from Michigan State University, East Lansing, Michigan, in 1985 and 1988, respectively. From 1988 to 1990, he was an assistant professor of computer and information science at New Jersey Institute of Technology, New Jersey. In 1990, he joined the faculty of the Department of Computer Science, National Tsing Hua University, Taiwan, where he is currently a professor. He served as the chair of the department from 2009 to 2012. His research interests include parallel and distributed processing and networked embedded systems.
\end{IEEEbiography}

\end{document}